\documentclass[letterpaper]{article} % DO NOT CHANGE THIS
\usepackage[submission]{aaai24}  % DO NOT CHANGE THIS
\usepackage{times}  % DO NOT CHANGE THIS
\usepackage{helvet}  % DO NOT CHANGE THIS
\usepackage{courier}  % DO NOT CHANGE THIS
\usepackage[hyphens]{url}  % DO NOT CHANGE THIS
\usepackage{amsmath}
\usepackage{amsthm}
\usepackage{amssymb}
\usepackage{graphicx} % DO NOT CHANGE THIS
\usepackage{natbib}  % DO NOT CHANGE THIS AND DO NOT ADD ANY OPTIONS TO IT
\usepackage{caption} % DO NOT CHANGE THIS AND DO NOT ADD ANY OPTIONS TO IT
\usepackage{algorithm}
\usepackage{algorithmic}

\usepackage{newfloat}
\usepackage{listings}
\DeclareCaptionStyle{ruled}{labelfont=normalfont,labelsep=colon,strut=off} % DO NOT CHANGE THIS
\floatstyle{ruled}
\newfloat{listing}{tb}{lst}{}
\floatname{listing}{Listing}
\newtheorem{example}{Example}
\newtheorem{theorem}{Theorem}
\newtheorem{lemma}[theorem]{Lemma}

\newtheorem{proposition}[theorem]{Proposition}
\newtheorem{definition}{\textbf{Definition}}

\usepackage{tikz,ifthen}
\usetikzlibrary{arrows,trees,backgrounds,automata,shapes,decorations,plotmarks,fit,calc,positioning,shadows,chains}

\definecolor{color0}{RGB}{240, 78, 64} %output
\definecolor{color4}{RGB}{60, 220, 125} %input
\definecolor{color6}{RGB}{120, 100, 200} %hidden
\definecolor{color7}{RGB}{107, 100, 200} %hidden

\definecolor{nnedgecolor}{RGB}{90,90,90}
\tikzstyle{every pin edge}=[<-,shorten <=1pt]
\tikzstyle{every path}=[draw=color7!50]
\tikzstyle{neuron}=[circle,fill=black!25,minimum size=17pt,inner sep=0pt]
\tikzstyle{input neuron}=[neuron, fill=color4]
\tikzstyle{output neuron}=[neuron, fill=color0]
\tikzstyle{hidden neuron}=[neuron, fill=color6]
\tikzstyle{annot} = [text width=4em, text centered]
\tikzstyle{nnedge} = [-{stealth},shorten >=0.1cm, shorten <=0.05cm,line 
width=0.8pt,nnedgecolor]
\tikzstyle{nnedge_t} = [-{dashed},shorten >=0.1cm, shorten <=0.05cm,line 
width=0.8pt,nnedgecolor]
\usetikzlibrary{calc}

\usepackage{multirow}

\title{Enumerating Safe Regions in Deep Neural Networks\\ with Provable Probabilistic Guarantees} 
\author{
    Luca Marzari\textsuperscript{\rm 1},
    Davide Corsi\textsuperscript{\rm 1},
    Enrico Marchesini\textsuperscript{\rm 2},
    Alessandro Farinelli\textsuperscript{\rm 1},
    Ferdinando Cicalese\textsuperscript{\rm 1}
}
\affiliations{
    \textsuperscript{\rm 1}Department of Computer Science, University of Verona, Italy\\
    \textsuperscript{\rm 2}Laboratory for Information \& Decision Systems, Massachusetts Institute of Technology, USA\\
     \{luca.marzari, davide.corsi, alessandro.farinelli, ferdinando.cicalese\}@univr.it\\
    emarche@mit.edu
}

\usepackage{bibentry}
\begin{document}
% 7 pages + references
\maketitle

\begin{abstract}
Identifying safe areas is a key point to guarantee trust for systems that are based on Deep Neural Networks (DNNs). To this end, we introduce the \textit{AllDNN-Verification} problem: given a safety property and a DNN, enumerate the set of all the regions of the property input domain which are safe, i.e., where the property does hold. Due to the \#P-hardness of the problem, we propose an efficient approximation method called \texttt{$\epsilon$-ProVe}.
Our approach exploits a controllable underestimation of the output reachable sets obtained via statistical prediction of tolerance limits, and can provide a tight —with provable probabilistic guarantees— lower estimate of the safe areas. Our empirical evaluation on different standard benchmarks shows the scalability and effectiveness of our method, offering valuable insights for this new type of verification of DNNs.

%%versione breve
% We consider the enumeration of all the safe regions for a trained Deep Neural Network and a given safety property. 
% Recent literature addresses the problem of counting safe regions, using Formal Verification tools, to estimate the probability of incurring in unsafe behaviors. However, counting provides no information on the distribution of safe areas, which is  crucial for many important tasks such as model ranking or selection. 

% To address this limitation, in this work, we introduce the \textit{AllDNN-Verification} problem, which aims to return a provably tight approximation of the set of all safe regions for a given safety property and a trained DNN; 
% Due to the \#P-Completeness of the problem, we propose an efficient approximation method called \texttt{$\epsilon$-ProVe}, which exploits the analysis of underestimated output reachable sets. Such reachable sets are obtained using statistical prediction of tolerance limits, hence providing a provable (probabilistic) lower bound of the safe areas. Our empirical evaluation on different standard FV benchmarks shows the scalability and effectiveness of our method, offering valuable insights for this new type of verification of DNNs.

\end{abstract}

\section{Introduction}
Deep Neural Networks (DNNs) have emerged as a groundbreaking technology revolutionizing various fields ranging from autonomous navigation \cite{navigation,Curriculum} to image classification \cite{image} and robotics for medical applications \cite{colon}. %Their ability to learn complex patterns from vast amounts of data has allowed them to tackle challenging tasks in several domains. %However, DNNs become more powerful and pervasive, safety concerns have become increasingly prominent. 
However, while DNNs can perform remarkably well in different scenarios, their reliance on massive data for training can lead to unexpected behaviors and vulnerabilities in real-world applications. In particular, DNNs are often considered "black-box" systems, meaning their internal representation is not fully transparent. A crucial DNNs weakness is the vulnerability to adversarial attacks \cite{adversarial, TACAS}, wherein small, imperceptible modifications to input data can lead to wrong and potentially catastrophic decisions when deployed. 

To this end, Formal Verification (FV) of DNNs \cite{Reluplex, Liu} holds great promise to provide assurances on the safety aspect of these functions before the actual deployment in real scenarios. In detail, the decision version of the \textit{DNN-Verification} problem takes as input a trained DNN $\mathcal{N}$ and a safety property, typically expressed as an input-output relationship for $\mathcal{N}$, and aims at determining whether there exists at least an input configuration which results in a violation of the safety property.
It is crucial to point out that due to the fact that the DNN's input is typically defined in a continuous domain any empirical evaluation of a safety property cannot rely on testing all the (infinitely many) possible input configurations.
In contrast, FV can provide provable assurances on the DNNs' safety aspect. However, despite the considerable advancements made by DNN-verifiers over the years \cite{Marabou,BetaCrown,Liu}, the binary result (\textit{safe} or \textit{unsafe}) provided by these tools is generally not sufficient to gain a comprehensive understanding of these functions. For instance, when comparing two neural networks and employing an FV tool that yields an \textit{unsafe} answer for both (i.e., indicating the existence of at least one violation point), we cannot distinguish whether one model exhibits only a small area of violation around the identified counterexample, while the other may have multiple and widespread violation areas.

To overcome this limitation, a quantitative variant of FV, asking for the number of violation points, has been proposed and analyzed, first in  \cite{Baluta} for the restricted class of Binarized Neural Networks (BNNs) and more recently in \cite{CountingProVe} for general DNNs. Following \cite{CountingProVe} we will henceforth refer to such counting problem as \textit{\#DNN-Verification}. 
Due to the \#P-hardness of the \textit{\#DNN-Verification}, both studies in  \cite{Baluta,CountingProVe}, focus on efficient approximate solutions, which allow  the resolution of large-scale real-world problems while providing provable (probabilistic) guarantees regarding the computed count. 

Solutions to the \textit{\#DNN-Verification} problem allow to estimate the probability that a DNN violates a given property but they do not provide information on the actual input configurations that are safe or violations for the property of interest. 

On the other hand, knowledge of the distribution of safe and unsafe areas in the input space is a key element to devise approaches that can enhance the safety of DNNs, e.g., by patching unsafe areas through re-training. 

To this aim, we introduce the 
\textit{AllDNN-Verification} problem 
%{\color{red} In the following paragraph, I am not sure we should  mention the approximate problem, but only the fact that we give approximate solutions. The introduction of the approximate problem is more a technical tool than an aim of our study. In fact the main theorem is now given with respect to the AllDNN problem. Unless we really wanna stress the compactness of the solution as a target}
, which corresponds to computing the set of all the areas that do not result in a violation for a given DNN and a safety property (i.e., enumerating all the safe areas of a property's input domain). The \textit{AllDNN-Verification} is at least as hard as \textit{\#DNN-Verification}, i.e., it is easily shown to be \#P-Hard. 
%{\color{blue}{Therefore, to address the challenges related to scalability and applicability to real-world scenarios, we propose an approximate version of the problem called \textit{$\epsilon$-Rectilinear Under-Approximation of safe areas for DNN ($\epsilon$-RUA-DNN)} which aims to return the set of all the safe regions using the cartesian products of hyperrectangles of size at least $\epsilon.$. The relaxation of this new problem has two main advantages: (i) we can use a simple representation of areas with easily analyzed geometric figures. (ii) it allows us to use interval analysis and thus propose solutions in continuous space without the need to discretize to count individual safe points within the domain. }}

Hence, we propose  \texttt{$\epsilon$-ProVe}, an approximation approach that provides 
 provable (probabilistic) guarantees on the returned areas. \texttt{$\epsilon$-ProVe} is built upon interval analysis of the DNN output reachable set and the iterative refinement approach \cite{reluval}, enabling efficient and reliable enumeration of safe areas.\footnote{We point out that the \textit{AllDNN-Verification} can also be defined to compute the set of unsafe regions. For better readability, we will only focus on safe regions. The definition and the solution proposed are directly derivable also when applied to unsafe areas.}

% In more detail, from the Lipshitz continuity property of DNNs \cite{reluval}, we can derive that DNNs are also uniformly continuous. 
% Therefore, by leveraging the \textit{Extreme Value Theorem} \cite{bolzano}, such DNN has both a minimum and a maximum value in a specific domain of interest. However, given the non-linear and non-convex nature of these function approximators, computing the exact minimum and maximum is impractical. For this reason, 

Notice that state-of-the-art FV methods typically propose an over-approximated output reachable set, thereby ensuring the soundness of the result. Nonetheless, the relaxation of the nonlinear activation functions employed to compute the over-approximate reachable set has non-negligible computational demands. In contrast, \texttt{$\epsilon$-ProVe} provides a scalable solution based on an underestimation of the output reachable set that exploits the \textit{Statistical Prediction of Tolerance Limits} \cite{wilks1942statistical, porter2019wilks}. In particular, we demonstrate how, with a confidence $\alpha$, our underestimation of the reachable set computed with $n$ random input configurations sampled from the initial property's domain $A$ is a correct output reachable set for at least a fraction $R$ of an indefinitely large further sample of points. Broadly speaking, this result tells us that if all the input configurations obtained in a random sample produce an output reachable set that does not violate the safety property (i.e., a \textit{safe} output reachable set) then, with probability $\alpha$, at least a subset of $A$ of size $R\cdot|A|$ is safe (i.e., $R$ is a lower bound on the safe rate in $A$ with confidence $\alpha$). 
%Crucially, by removing from the starting domain the provable safe regions obtained by our approximation and adding a percentage of areas equal to the $1-R$, that we could only obtain by an exact enumeration method, it is possible to get a provable probabilistic upper bound of the unsafe areas in the same domain of interest. 
In summary, the main contributions of this paper are the following:
\begin{itemize}
    \item We initiate the
%    To the best of our knowledge, this is the first 
study of the 
%to consider the 
\textit{AllDNN-verification} problem, the enumeration version of the DNN-Verification
%problem.
    \item Due to the \#P-hardness of the problem, we propose \texttt{$\epsilon$-ProVe} a novel %polynomial 
    approximation method to obtain a provable (probabilistic) lower bound of the safe zones within a given property’s domain.
    % \item {\color{blue}Due to the \#P-hardness of the problem, we propose a relaxation of \textit{AllDNN-verification}, namely \textit{$\epsilon$-RUA-DNN}, which requires the set of safe regions exploiting $\epsilon$-bounded hyperrectangles.}
    % \item {\color{blue}We present \texttt{$\epsilon$-ProVe} a novel polynomial approximation method to solve \textit{$\epsilon$-RUA-DNN} and obtain a provable (probabilistic) lower bound of the safe regions within a given property's domain. }
    \item We evaluate our approach on FV standard benchmarks, showing that \texttt{$\epsilon$-ProVe} is scalable and effective for real-world scenarios.
\end{itemize}

\section{Preliminaries}

In this section, we discuss existing FV approaches and related key concepts on which our approach is based. 
In contrast to the standard robustness and adversarial attack  literature \cite{robustness,advAttack,zhang2022branch}, FV of DNNs seeks formal guarantees on the safety aspect of the neural network given a specific input domain of interest. Broadly speaking, if a DNN-Verification tool states that the region is provably verified, this implies that there are no adversarial examples  -- violation points -- in that region. 
 We recall in the next section the formal definition of the satisfiability problem for \textit{DNN-Verification} \cite{Reluplex}.

\subsection{DNN-Verification}

In the \textit{DNN-Verification}, we have as input a tuple $\mathcal{T}=\langle\mathcal{N}, \mathcal{P}, \mathcal{Q}\rangle$, where $\mathcal{N}$ is a DNN, $\mathcal{P}$ is precondition on the input, and $\mathcal{Q}$ a postcondition on the output. In particular, $\mathcal{P}$ denotes a particular input domain or region for which we require a particular postcondition $\mathcal{Q}$ to hold on the output of $\mathcal{N}$.  
Since we are interested in discovering a possible counterexample, $\mathcal{Q}$ typically encodes the negation of the desired behavior for $\mathcal{N}$.
Hence, the possible outcomes are $\texttt{SAT}$ if there exists an input configuration that lies in the input domain of interest, satisfying the predicate $\mathcal{P}$, and for which the DNN satisfies the postcondition $\mathcal{Q}$, i.e., at least one violation exists in the considered area, $\texttt{UNSAT}$ otherwise. 

To provide the reader with a better intuition on the \textit{DNN-Verification} problem, we discuss a toy example.
\begin{example}(DNN-Verification)
 Suppose we want to verify that for the toy DNN $\mathcal{N}$ depicted in Fig. \ref{fig:toyDNN} given an input vector $x=(x_1, x_2) \in [0,1]\times[0,1]$, the resulting output should always be $y\geq0$. We define 
$\mathcal{P}$ as the predicate  on the input vector $x = (x_1, x_2)$ which is true  iff  $x \in [0,1]\times [0,1]$, and $\mathcal{Q}$ as the predicate on the output $y$ which is true iff $y = \mathcal{N}(x) < 0,$ that is, we set $\mathcal{Q}$ to be the negation of our desired property. As reported in Fig. \ref{fig:toyDNN}, given the vector $x=(1,0)$ we obtain $y<0$, hence the verification tool returns a $\texttt{SAT}$ answer, meaning that a specific counterexample exists and thus the original safety property does not hold.
\begin{figure}[h!]
	\begin{center}
		\scalebox{0.8} {
			\def\layersep{2.0cm}
			\begin{tikzpicture}[shorten >=1pt,->,draw=black!50, node
				distance=\layersep,font=\footnotesize]
				
				\node[input neuron] (I-1) at (0,-1) {$x_1$};
				\node[input neuron] (I-2) at (0,-2.5) {$x_2$};

                    % vector input
				\node[left=-0.05cm of I-1] (b1) {$[1]$};
				\node[left=-0.05cm of I-2] (b2) {$[0]$};
				
				\node[hidden neuron] (H-1) at (\layersep,-1) {$h_1$};
				\node[hidden neuron] (H-2) at (\layersep,-2.5) {$h_1$};
				
				\node[hidden neuron] (H-3) at (2*\layersep,-1) {$h_1^a$};
				\node[hidden neuron] (H-4) at (2*\layersep,-2.5) {$h_2^a$};
				
				\node[output neuron] at (3*\layersep, -1.75) (O-1) {$y$};
				
				% Connect every node in the hidden layer with the output layer
				\draw[nnedge] (I-1) --node[above] {$4$} (H-1);
				\draw[nnedge] (I-1) --node[above, pos=0.3] {$-2$} (H-2);
				\draw[nnedge] (I-2) --node[below, pos=0.3] {$-1$} (H-1);
				\draw[nnedge] (I-2) --node[below] {$3$} (H-2);
				
				\draw[nnedge] (H-1) --node[above] {ReLU} (H-3);
				\draw[nnedge] (H-2) --node[below] {ReLU} (H-4);
				
				\draw[nnedge] (H-3) --node[above] {$-1$} (O-1);
				\draw[nnedge] (H-4) --node[below] {$7$} (O-1);

				% result first prop
				\node[below=0.05cm of H-1] (b1) {$[+4]$};
				\node[below=0.05cm of H-2] (b2) {$[-2]$};

                     % Biases
				\node[below=0.05cm of H-3] (b1) {$[+4]$};
				\node[below=0.05cm of H-4] (b2) {$[0]$};

                    \node[right=0.05cm of O-1] (b1) {$[-4]$};

				% Annotate the layers
				\node[annot,above of=H-1, node distance=0.8cm] (hl1) {Weighted
					sum};
				\node[annot,above of=H-3, node distance=0.8cm] (hl2) {Activated Layer };
				\node[annot,left of=hl1] {Input };
				\node[annot,right of=hl2] {Output };
			\end{tikzpicture}
		} 
		
	\end{center}
    \caption{A counterexample for a toy \textit{DNN-Verification} problem.}
    \label{fig:toyDNN}
    \vspace{-3mm}
\end{figure}
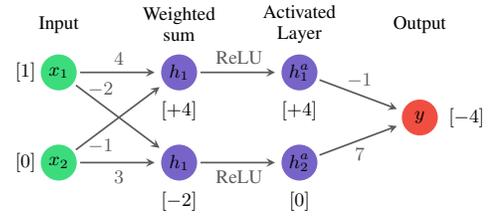
\end{example}

\subsection{\#DNN-Verification}
Despite the provable guarantees and the advancement that formal verification tools have shown in recent years \cite{Liu,Marabou,BetaCrown,zhang2022general}, the binary nature of the result of the \textit{DNN-Verification} problem may hide additional information about the safety aspect of the DNNs. To address this limitation in \cite{CountingProVe} the authors introduce the 
\textit{\#DNN-Verification}, i.e., the extension of the decision problem to its counting version. In this problem, the input is the same as the decision version, but we denote as  $\Gamma(\mathcal{T})$ the set of \textit{all} the input configurations for $\mathcal{N}$ satisfying the property defined by $\mathcal{P}$ and $\mathcal{Q},$
i.e.
\begin{equation} \label{Gamma} 
\Gamma(\mathcal{T}) = \Bigg\{ x \; \big\vert \; \mathcal{P}(x) \wedge \mathcal{Q}(\mathcal{N}(x)) \Bigg\} 
\end{equation}
Then, the 
\textit{\#DNN-Verification} consists of computing $\vert \Gamma(\mathcal{T})\vert$.

\begin{figure}[t]
    \centering
    \includegraphics[width=0.9\linewidth]{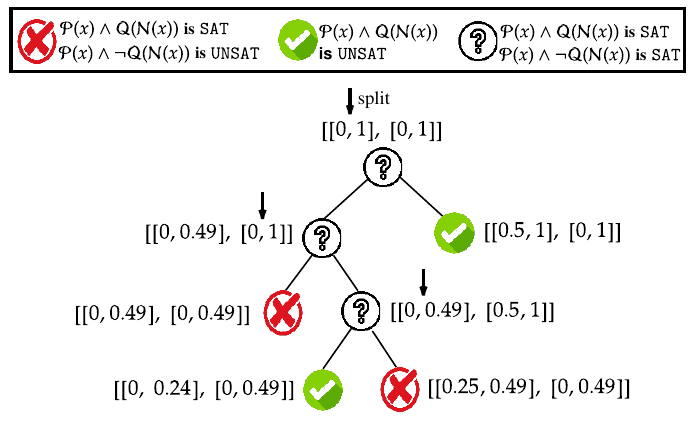}
    \caption{Explanatory image execution of exact count for a particular $\mathcal{N}$ and safety property.}
    \label{fig:exactCount}
\end{figure}

The approach (reported in Fig.\ref{fig:exactCount}) solves the problem in a sound and complete fashion where any state-of-the-art FV tool for the decision problem can be employed to check each node of the Branch-and-Bound \cite{BaB} tree recursively. In detail, each node produces a partition of the input space into two equal parts as long as it contains both a point that violates the property and a point that satisfies it. The leaves of this recursion tree procedure correspond to partitioning the input space into parts where we have either complete violations or safety. Hence, the provable count of the safe areas is easily computable by summing up the cardinality of the subinput spaces in the leaves that present complete safety. Since our setting is in the continuum, the number of points in any non-empty set is infinite. Hence, we consider the cardinality as a proxy for the volume of the corresponding set. Nonetheless, if we assume some discretization of the space (to the maximum resolution allowed by the machine precision), $\Gamma(\mathcal{T})$ becomes a finite countable set.

Clearly, by using this method, it is possible to exactly count (and even to enumerate) the safe points. 
However, due to the necessity of solving a \textit{DNN-Verification} instance at each node (an intractable problem that might require exponential time), this approach becomes soon unfeasible and struggles to scale on real-world scenarios. 
In fact, it turns out that under standard complexity assumption, no efficient and scalable approach can return the exact set of areas in which a DNN is provably safe (as detailed in the next section). 

To address this concern, after formally defining the \textit{AllDNN-Verification} problem and its complexity, we propose first of all a relaxation of the problem, and subsequently, an approximate method that exploits the analysis of underestimated output reachable sets obtained using statistical prediction of tolerance limits \cite{wilks1942statistical, porter2019wilks} and provides a tight underapproximation of the safe areas with strong probabilistic guarantees.

\section{The \textit{AllDNN-Verification} Problem}
The \textit{AllDNN-Verification} problem asks for the set of all the safe points for a particular tuple $\langle\mathcal{N}, \mathcal{P}, \mathcal{Q}\rangle$.
Formally:
\begin{definition}[\textit{AllDNN-Verification Problem}]
\phantom{abc}

{\bf Input}: A tuple $\mathcal{T}=\langle\mathcal{N}, \mathcal{P}, \mathcal{Q}\rangle$. 

{\bf Output}: the set of safe points $\Gamma(\mathcal{T}),$ as given in (\ref{Gamma}).
\end{definition}

%with 
%\[ \Gamma(\mathcal{T}) = \Bigg\{ r_i \; \big\vert \; \exists x \in r_i \;s.t.\; \mathcal{P}(x) \wedge \mathcal{Q}(\mathcal{N}(x)) \Bigg\} \] 
%and where $r_i \cap r_j = \emptyset \;\; \forall i \neq j$.

%Hence, $\Gamma(\mathcal{T})$ is the set of the regions 
%%composed of hyperrectangles 
%for which exists at least an input configuration in the area that satisfies the property defined by $\mathcal{P}$ and $\mathcal{Q}$, and for any pair of areas, the intersection is empty, i.e., all the disjoint areas. In particular, 
Considering the example of Fig. \ref{fig:exactCount} solving the \textit{AllDNN-Verification} problem for the safe areas consists in returning the set:
$\displaystyle{\Gamma = \Bigg\{\big[[0.5, 1] \times [0, 1]\big] \cup  \big[[0, 0.24] \times [0, 0.49]\big]\Bigg\}}.$
%%\Bigg\{\underbrace{\big[[0, 0.49], [0, 0.49]\big]}_{r_0}, \underbrace{\big[[0.25, 0.49], [0, 0.49]\big]}_{r_1}\Bigg\}
%where for every $r_i \in \Gamma$ with $i=\{0,1\}$ there exists at least a violation point\footnote{Notice that violations point can be infinitely many since we are in a continuous space} for the tuple $\mathcal{T}$.

\subsection{Hardness of \textit{AllDNN-Verification}}

From the \#P-hardness of the \textit{\#DNN-Verification} problem proved in  \cite{CountingProVe} and the fact that exact enumeration also provides exact counting it immediately follows that  the \textit{AllDNN-Verification} is  \textit{\#P-hard}, which essentially states that no polynomial algorithm is expected to exists for the \textit{AllDNN-Verification} problem.

\section{$\epsilon$-ProVe: a Provable (Probabilistic) Approach}
In view of the structural scalability issue of any solution to the \textit{AllDNN-Verification} problem, due to its \#P-hardness, we propose to resort to an approximate solution. More precisely, we define the following approximate version of  the \textit{AllDNN-Verification} problem:

\begin{definition}[\textit{$\epsilon$-Rectilinear Under-Approximation of safe areas for DNN ($\epsilon$-RUA-DNN)}]
\phantom{abc}

{\bf Input}: A tuple $\mathcal{T}=\langle\mathcal{N}, \mathcal{P}, \mathcal{Q}\rangle$. 

{\bf Output}: a family ${\cal R}= \{r_1, \dots, r_m\}$ of disjoint rectilinear $\epsilon$-bounded hyperrectangles such that $\bigcup_i r_i \subseteq \Gamma({\cal T})$ and 
$|\Gamma({\cal T}) \setminus \bigcup_i r_i |$ is minimum.
\end{definition}

A {\em rectilinear $\epsilon$-bounded hyperrectangle} is defined as the cartesian product of intervals of size at least $\epsilon.$
%Moreover, for $\gamma > 0,$  we say that a rectilinear hyperrectangle $r = \times_i [\ell_i, u_i]$ is {\em $\gamma$-aligned}
%if for each $i,$ both extremes $\ell_i$ and $u_i$ are a  multiple of $\gamma.$ 
Moreover, for $\epsilon > 0,$  we say that a rectilinear hyperrectangle $r = \times_i [\ell_i, u_i]$ is {\em $\epsilon$-aligned}
if for each $i,$ both extremes $\ell_i$ and $u_i$ are a  multiple of $\epsilon.$ 

The rationale behind this new formulation of the problem is twofold: on the one hand, we are relaxing the request for the exact enumeration of safe points---in fact, as argued in \cite{karp1985monte} due to the \#P-hardness proof (from \#3-SAT), even guaranteeing a constant approximation to $|\Gamma({\cal T})|$  by a deterministic polynomial time algorithm is not possible unless $P = NP$; 
on the other hand, we are requiring that the output is more concisely representable by means of hyperrectangles of {\em some significant} size. 

Note that for $\epsilon \to 0$, \textit{$\epsilon$-RUA-DNN} and \textit{AllDNN-Verification} become the same problem. More generally, whenever the solution $\Gamma({\cal T})$ to an instance ${\cal T}$ of \textit{AllDNN-Verification} can be partitioned into a collection of rectilinear $\epsilon$-bounded hyperrectangles, $\Gamma({\cal T})$ can be attained by an optimal solution for  the \textit{$\epsilon$-RUA-DNN}.  This allows us to tackle the \textit{AllDNN-Verification} problem via an efficient approach with strong probabilistic approximation guarantees to solve  the \textit{$\epsilon$-RUA-DNN} problem.

%with 
%\[ \Gamma(\mathcal{T}) = \Bigg\{ r_i \; \big\vert \; \exists x \in r_i \;s.t.\; \mathcal{P}(x) \wedge \mathcal{Q}(\mathcal{N}(x)) \Bigg\} \] 
%and where $r_i \cap r_j = \emptyset \;\; \forall i \neq j$.

%Hence, $\Gamma(\mathcal{T})$ is the set of the regions 
%%composed of hyperrectangles 
%for which exists at least an input configuration in the area that satisfies the property defined by $\mathcal{P}$ and $\mathcal{Q}$, and for any pair of areas, the intersection is empty, i.e., all the disjoint areas. In particular, 

\begin{figure}[t]
    \centering
    \includegraphics[width=\linewidth]{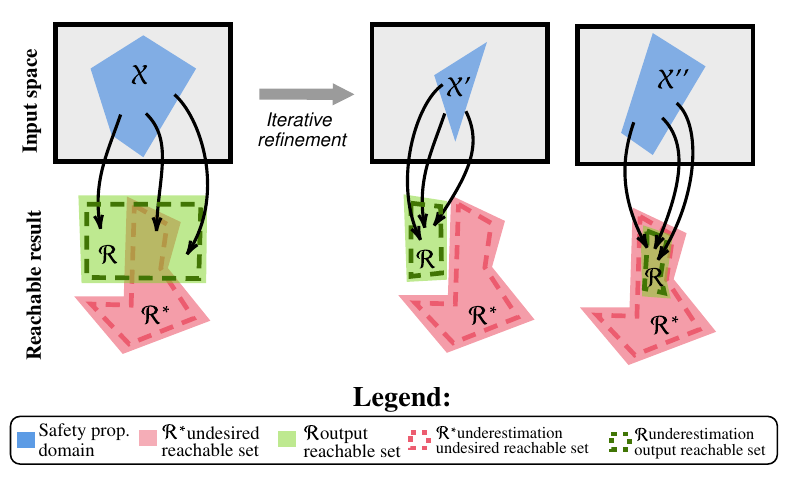}
    \caption{Explanatory image of how to exploit reachable set result for solving the \textit{AllDNN-Verification} problem. }
    \label{fig:reachable}
\end{figure}

Our method is based  on two main concepts: the analysis of an underestimated output reachable set with probabilistic guarantees and the \textit{iterative refinement} approach \cite{reluval}. 
In particular, in Fig. \ref{fig:reachable} we report a schematic representation of the approach that can be set up through reachable set analysis.
Let us consider a possible domain for the safety property, i.e., the polygon highlighted in light blue in the upper left corner of Fig. \ref{fig:reachable}.  

Suppose that the undesired output reachable set is the one highlighted in red called $\mathcal{R^*}$ in the bottom left part of the image, i.e., this set describes all the unsafe outcomes the DNN should never output starting from $\mathcal{X}$. Hence, in order to formally verify that the network respects the desired safety property, the output reachable set computed from the domain of the property (i.e., the green $\mathcal{R}$ area of the left side of the image) should have an empty intersection with the undesired reachable set (the red one). If this condition is not respected, as, e.g., in the left part of the figure, then there exists at least an input configuration for which the property is not respected. 

%Even though this approach provides a way to provably verify the safety of our DNN (and it is used in many state-of-the-art DNN-verification tools), it does not give us any information about how many or what are, the (un)safe areas in the domain considered by the property. 
To find all the portions of the property's domain where either the undesired reachable set and the output reachable set are disjoint, i.e., $\mathcal{R^*} \bigcap \mathcal{R} = \emptyset$, 
%i.e., compute all the provable safe regions of the property domain, 
or, dually, discover the unsafe areas where the condition $\mathcal{R} \subseteq \mathcal{R^*}$ holds (as shown in the right part of Fig. \ref{fig:reachable}) we can exploit the \textit{iterative refinement} approach \cite{reluval}. 
However, given the nonlinear nature of DNNs, computing the exact output reachable set is infeasible. To address this issue, the reachable set is typically over-approximated, thereby ensuring the soundness of the result. In this vein, \cite{yang2022neural} proposed an enumeration approach based on an over-approximation of the reachable set to compute the set of unsafe regions in the property's input domain. Still, the relaxation of the nonlinear activation functions used to compute the over-approximated reachable set can be computationally demanding. In contrast, we propose a computationally efficient solution that uses underestimation of the reachable set and constructs approximate solutions for the \textit{$\epsilon$-RUA-DNN} problem with strong probabilistic guarantees.

\subsection{Probabilistic Reachable Set}

Given the 
%considerations on the
complexity of computing the exact minimum and maximum of the function computed by a DNN, we propose to approximate the output reachable set using a statistical approach known as \textit{Statistical Prediction of Tolerance Limits} \cite{wilks1942statistical, porter2019wilks}.

%In detail, tolerance limits are statistical measures used to establish the range within which a measurement on all the members of a proportion is expected to fall given a first prediction based on the measurement of $n$ sampled individuals. Given a specific domain of interest, we can consider a possible output reachable set $X$ as a continuous random variable and assume that there exists some probability distribution function $f(x)$ of values of $X$, such that
%$\int_a^b f(x)\;dx$ is the probability that $a < X < b, \; \forall a,b \in \mathbb{R}$. The integral is used to compute the area under a probability density function curve, which represents a specific proportion of the population that falls within a certain range. 

We use a Monte Carlo sampling approach: for an appropriately chosen $n$, we sample $n$ input points and take the smallest and the greatest value achieved in the output node as the lower and the upper extreme of our probabilistic estimate of the reachable set. The choice of the sample size is based on the results of \cite{wilks1942statistical} that allow us to choose $n$ in order to achieve a given desired guarantee on the probability $\alpha$ that our estimate of the output reachable set holds for at least a fixed (chosen) fraction $R$ of a further possibly infinitely large sample of inputs. 
Crucially, this statistical result does not require any knowledge of the probability distribution governing our function of interest and thus also applies to general DNNs.
Stated in terms more directly applicable to our setting, the main result of \cite{wilks1942statistical} is as follows.
%:
%$P\Big( \int_L^U f(x)\;dx \geq R\Big) = \alpha$,
%%\[P\Big( \int_L^U f(x)\;dx \geq R\Big) = \alpha\]
%where $L$ and $U$ are the lower and the upper bound of the output reachable set estimated.
%In fact, in the next section, we show how for the computation of (un)safe areas, it is enough to check one part of the estimated output reachable set, checking whether the reachable set is positive or not. This allows us to leverage the solution based on the multinomial distribution law proposed in \cite{wilks1942statistical} for the \textit{one tolerance limit} problem, i.e., the following result:

\begin{lemma}\label{lemma_Wilks}
    For any  $R \in (0,1)$ and integer $n$, given a sample of $n$ values from a (continuous) set $X$ the probability that for at least a fraction $R$ of the values in a further possibly infinite sequence of samples from  $X$ are all not smaller (respectively larger) than the minimum value  (resp.\ maximum value) estimated with the first $n$ samples is given by the $\alpha$ satisfying the following equation
    \begin{equation} \label{eq:Wilks}
        n \cdot \int_R^1 x^{n-1}\;dx = (1 - R^n) = \alpha
    \end{equation}
    
\end{lemma}

\subsection{Computation of Safe Regions}
We are now ready to give a detailed account of our algorithm  \texttt{$\epsilon$-ProVe}. 

\begin{figure}[t]
	\begin{center}
		\scalebox{.8} {
			\def\layersep{2.0cm}
			\begin{tikzpicture}[shorten >=1pt,->,draw=black!50, node
				distance=\layersep,font=\footnotesize]
				
				\node[input neuron] (I-1) at (0,-1) {$x_1$};
				\node[input neuron] (I-2) at (0,-2.5) {$x_2$};

                    % vector input
				\node[left=-0.05cm of I-1] (b1) {$[1, 0]$};
				\node[left=-0.05cm of I-2] (b2) {$[0, 1]$};
				
				\node[hidden neuron] (H-1) at (\layersep,-1) {$h_1$};
				\node[hidden neuron] (H-2) at (\layersep,-2.5) {$h_1$};
				
				\node[hidden neuron] (H-3) at (2*\layersep,-1) {$h_1^a$};
				\node[hidden neuron] (H-4) at (2*\layersep,-2.5) {$h_2^a$};
				
				\node[output neuron] at (3*\layersep, -1) (O-1) {$y_1$};

                \node[output neuron] at (3*\layersep, -2.5) (O-2) {$y_2$};

                 \node[output neuron] at (4*\layersep, -1.75) (O-3) {$y^*$};

				% Connect every node in the hidden layer with the output layer
				\draw[nnedge] (I-1) --node[above] {$4$} (H-1);
				\draw[nnedge] (I-1) --node[above, pos=0.3] {$-2$} (H-2);
				\draw[nnedge] (I-2) --node[below, pos=0.3] {$-1$} (H-1);
				\draw[nnedge] (I-2) --node[below] {$3$} (H-2);
				
				\draw[nnedge] (H-1) --node[above] {ReLU} (H-3);
				\draw[nnedge] (H-2) --node[below] {ReLU} (H-4);

				\draw[nnedge] (H-3) --node[above] {$-1$} (O-1);
                \draw[nnedge] (H-3) --node[above, pos=0.3] {$1$} (O-2);
				\draw[nnedge] (H-4) --node[below] {$2$} (O-2);
                \draw[nnedge] (H-4) --node[below, pos=0.3] {$0$} (O-1);

                \draw[nnedge] (O-1) --node[above] {$-1$} (O-3);
				\draw[nnedge] (O-2) --node[below] {$1$} (O-3);

				% result first prop
				\node[below=0.05cm of H-1] (b1) {$[+4, -1]$};
				\node[below=0.05cm of H-2] (b2) {$[-2, +3]$};

                     % Biases
				\node[below=0.05cm of H-3] (b1) {$[+4, 0]$};
				\node[below=0.05cm of H-4] (b2) {$[0, +3]$};

                \node[below=0.05cm of O-1] (b1) {$[-4, +1]$};

                \node[below=0.05cm of O-2] (b2) {$[+4, +6]$};

                \node[right=0.05cm of O-3] (b2) {$[8, 5]$};

                \node[below=0.05cm of O-3] (b2) {$\mathcal{R} = [5, 8]$};

				% Annotate the layers
				\node[annot,above of=H-1, node distance=0.8cm] (hl1) {Weighted
					sum};
				\node[annot,above of=H-3, node distance=0.8cm] (hl2) {Activated Layer };
				\node[annot,left of=hl1] {Input };
				\node[annot,right of=hl2] {Output };
                \node[annot,above of=O-3, node distance=1.5cm] (o3) {New output layer };
			\end{tikzpicture}
		} 
		
	\end{center}
    \caption{Example of computation single reachable set for a DNN with two outputs.}
    \label{fig:SingleRechableSet}
    \vspace{-3mm}
\end{figure}
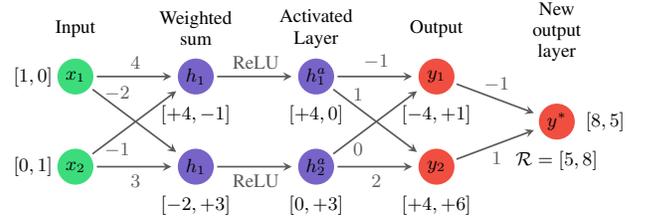

Our approximation is based on the analysis of an underestimated output reachable set obtained by sampling a set of $n$ points $P_A$ from a domain of interest $A$. 
We start by observing that it is possible to assume, without loss of generality, that the network has a single output node on whose reachable set we can 
verify the desired property \cite{Liu}. For networks not satisfying this assumption, we can enforce it by adding one layer. 
%we reduce the computation of can assume that the For all practical purposes, instead of computing one single reachable set for each output node, it is always possible to obtain a single reachable set given a specific condition to check, no matter the original cardinality of the output space \cite{Liu}. 
For example, consider the network in the example of Fig. \ref{fig:SingleRechableSet} and suppose we are interested in knowing if, for a given input configuration in a domain $A = [0,1]\times[0,1]$, the output $y_1$ is always less than $y_2$. By adding a new output layer with a single node $y^*$ connected to  $y_1$ by weight $-1$ and to $y_2$ with weight $1$ the condition required reduces to check that all the values in the reachable set for $y^*$ are positive.  

In general, from the analysis of the underestimated reachable set of the output node computed as $\mathcal{R} = [min_i \, y_i, max_i \, y_i],$
%\;\forall i=1,\dots,n$ 
we can obtain one of these three conditions:
\begin{equation} \label{safe-unsafe-cases}
    \begin{cases}
        \text{$A$ is unsafe} \quad &\text{upper bound of } \mathcal{R} < 0  \\
        \text{$A$ is safe} \quad &\text{lower bound of } \mathcal{R} \geq 0 \\
        unknown\quad &\text{otherwise}
    \end{cases}
\end{equation}

With reference to the toy example in Figure \ref{fig:SingleRechableSet}, assuming we sample only $n=2$ input configurations,  $(1,0)$ and $(0,1)$ which when propagated through $\mathcal{N}$ produce as a result in the new output layer the vector $y^* = [8, 5]$. This results in the estimated reachable set $\mathcal{R} = [5, 8]$. Since the lower bound of this interval is positive, we conclude that the region $A$, under consideration, is completely safe. 

To confirm the correctness of our construction, we can check the partial values of the original output layer and notice that no input generates $y_1 \geq y_2$. Specifically, if all inputs result in $y_1 \geq y_2$ (violating the specification we are trying to verify), then the reachable set must have, by construction, a negative upper bound, leading to the correct conclusion that the area is unsafe. On the other hand, if only some inputs produce $y_1 \geq y_2$, then we obtain a reachable set with a negative lower bound and a positive upper bound, thus we cannot state whether the area is unsafe or not, and we should proceed with an interval refinement process. Hence, this approach allows us to obtain the situations shown to the right of Fig. \ref{fig:reachable}, i.e., where the reachable set is either completely positive ($A$ safe) or completely negative ($A$ unsafe).

We present the complete pipeline of \texttt{$\epsilon$-ProVe} in Algorithm \ref{alg:eProVe}. Our approach receives as input a standard tuple for the \textit{DNN-Verification} and creates the augmented DNN $\mathcal{N}'$ (line 3) following the intuitions provided above. Moreover, we initialize respectively the set of safe regions as an empty set and the unchecked regions as the entire domain of the safety specification encoded in $\mathcal{P}$ (line 5). 
Inside the loop (line 6), our approximation iteratively considers one area $A$ at the time and begins computing the reachable set, as shown above. We proceed with the analysis of the interval computed, where in case we obtain a positive reachable set, i.e., the lower bound is positive (lines 9-10), then the area under consideration is deemed as safe and stored in the set of safe regions we are enumerating. On the other hand, if the interval is negative, that is, the upper bound is negative, we add the area into the unsafe regions and proceed (lines 11-12). Finally, if we are not in any of these cases, we cannot assert any conclusions about the nature of the region we are checking, and therefore, we must proceed with splitting the area according to the heuristic we prefer (lines 13-14).

The loop ends when either we have checked all areas of the domain of interest or we have reached the $\epsilon$-precision on the iterative refinement. In detail, given the continuous nature of the domain, it is always possible to split an interval into two subparts, that is, the process could continue indefinitely in the worst case. For this reason, as is the case of other state-of-the-art FV methods that are based on this approach, we use a parameter to decide when to stop the process. This does not affect the correctness of the output since our goal is to (tightly) underapproximate
%returns a provable probabilistic lower bound of 
the safe regions, and thus, in case the $\epsilon$-precision is reached, the area under consideration would not be considered in the set that the algorithm returns, thus preserving the correctness of the result. Although the level of precision can be set arbitrarily, it does have an effect on the performance of the method. In the supplementary material, we discuss the impact that different heuristics and hyperparameter settings have on the resulting approximation. %and a computational time and space analysis of our approach.

%This approach allows us to acquire a conservative estimation of the reachable set by randomly sampling points from a theoretically infinite space, resulting in a low likelihood of encountering the actual minimum or maximum values, which represent the true lower and upper bounds, respectively. However, we show the following section, that following the consideration of \cite{wilks1942statistical}, given a precise confidence in the answer and the lower bound that we want is possible to estimate the number of sample required to achieve that precision.

\begin{algorithm}[t]
\caption{\texttt{$\epsilon$-ProVe}}\label{alg:eProVe}
\begin{algorithmic}[1]
\small
\STATE \textbf{Input:} $\mathcal{T} =\langle\mathcal{N},\mathcal{P}, \mathcal{Q}\rangle$, $n$ ($\#$ of samples to compute $\mathcal{R}$), $\epsilon$-precision desired
\STATE \textbf{Output: } set of safe and unsafe regions in $\mathcal{P}$.
\vspace{0.2cm}

\STATE $\mathcal{N'} \gets \texttt{CreateAugmentedDNN}(\mathcal{N},\mathcal{P}, \mathcal{Q})$
\STATE safe\_regions $\gets \emptyset$; unsafe\_regions $\gets \emptyset$ 
\STATE unknown $\gets \texttt{GetDomain}(\mathcal{P})$
\WHILE{(unknown $\neq \emptyset$) or $(\epsilon$-precision not reached)}
    \STATE $A\gets \texttt{GetAreaToVerify}$(unknown)
    \STATE  $\mathcal{R}_A \gets \texttt{ComputeReachableSet}(\mathcal{N}',\; A,\; n)$
    \IF{$\texttt{lower}(\mathcal{R}_A) \geq 0$}
        \STATE safe\_regions $\gets$ safe\_regions $\cup\;\{A\}$ 
    \ELSIF{$\texttt{upper}(\mathcal{R}_A) < 0$}
        %\STATE continue
        \STATE unsafe\_regions $\gets$ unsafe\_regions $\cup\;\{A\}$
    \ELSE
        \STATE unknown $\gets$ unknown $\cup \;\texttt{IntervalRefinement}(A)$   
    \ENDIF   
\ENDWHILE

\STATE \textbf{return} safe\_regions, unsafe\_regions
\end{algorithmic}

\end{algorithm}

\subsection{Theoretical Guarantees}

In this section, we analyze the theoretical guarantees that our approach can provide. %For the sake of simplifying the presentation, in this section we take the perspective of wanting to identifying the safe areas---which is coherent with the presentation of the pseudocode. Clearly, the same results apply {\em mutatis mutandis} if the approach is used to estimate the violation areas.
We assume that the {\tt IntervalRefinement} procedure consists of iteratively choosing one of the dimensions of the input domain and splitting the area into two halves of equal size as in \cite{reluval}. The theoretical guarantee easily extends to any other heuristic provided that each split produces two parts both at least a fixed constant fraction $\beta$ of the subdivided area.
%\leq 1/2$ of the original dimension, thus obtaining splits in fractions $(\beta,\; 1-\beta)$.} %along the middle point of the interval that defines it in the chosen dimension.
%This immediately implies that the areas output by the algorithms are 
%$\gamma$-aligned rectilinear hyperrectangles with $\gamma = 2^{-g}$ where $g$ is the maximum number of times a split is performed in the same dimension. 
Moreover, we assume that reaching the $\epsilon$ precision is implemented as testing that the area has reached size $\epsilon^d,$, i.e., it is the cartesian product of $d$ intervals of size $\epsilon.$  It follows that, by definition, the areas output by \texttt{$\epsilon$-ProVe} are $\epsilon$-bounded and $\epsilon$-aligned.

The following proposition is the basis of the approximation guarantee (in terms of the size of the safe area returned) on the solution output by \texttt{$\epsilon$-ProVe} on an instance of  the \textit{$\epsilon$-RUA-DNN} problem.

\begin{proposition}\label{prop2}
Fix a real number $\epsilon > 0,$ an integer $k \geq 3,$ and a real $\gamma > k \epsilon.$ 
Let $\cal T$ be an instance of the $\epsilon$-RUA-DNN problem. Then for any solution ${\cal R} = \{r_1, \dots, r_m\}$ such that for each $i=1, \dots, m, $ $r_i$ is $\gamma$-bounded, there is a solution 
${\cal R}^{(\epsilon)} = \{r_1^{(\epsilon)}, \dots, r_m^{(\epsilon)}\}$ such that each $r_i^{(\epsilon)}$ is $\epsilon$-aligned and $||{\cal R}^{(\epsilon)}|| \geq \left(\frac{k-2}{k}\right)^d ||{\cal R}||,$ where
$d$ is the number of dimensions of the input space, and for every solution ${\cal R}'$, $||{\cal R}'|| = | \cup_i r_i|$ is the total area covered by the hyperrectangles in ${\cal R}'$. 
\end{proposition}

The result is obtained by applying the following lemma to each hyperrectangle of the solution ${\cal R}.$

\begin{lemma}\label{lemma3}
Fix a real number $\epsilon > 0$ and an integer $k \geq 3.$ For any $\gamma > k \epsilon$ and any $\gamma$-bounded rectilinear hyperrectangle 
$r \subseteq \mathbb{R}^d,$ 
there is an $\epsilon$-aligned rectilinear hyperrectangle $r^{(\epsilon)}$ such that: (i) $r^{(\epsilon)} \subseteq r$; and 
(ii) $|r^{(\epsilon)}| \geq \left( \frac{k-2}{k}\right)^d |r|.$
\end{lemma}

\begin{figure}[h!]
    \vspace{-5mm}
    \centering
    \includegraphics[width=0.9\linewidth]{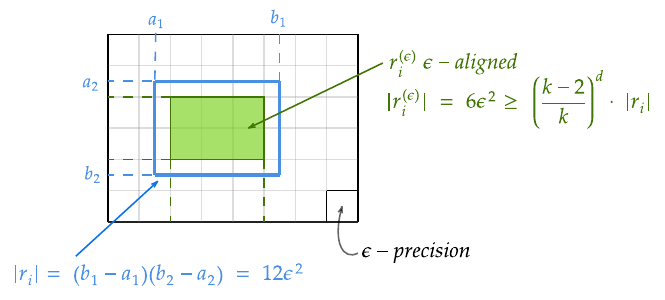}
    \vspace{-4mm}
    \caption{An example of applying Lemma \ref{lemma3} with $k=3$.}
    \label{fig:prop}
\end{figure}

Fig. \ref{fig:prop} gives a pictorial explanation of the lemma. In the example shown, $k = 3$ and the parameter $\epsilon$ is the unit of the grid, which we can imagine superimposed to the bidimensional ($d=2$) space in which the hyperrectangles live. Hence $\gamma = 3\epsilon.$ The non-$\epsilon$-aligned $r_i$ (depicted in blue) is $\gamma$-bounded, since its width $w= (b_1-a_1) = 4\epsilon$ and its height $h = (b_2-a_2) = 3\epsilon$ are both $\geq 3\epsilon.$ Hence, it covers completely at least $w-2$ columns and $h-2$ rows of the grid. These rows and columns define the green $\epsilon$-aligned hyperrectangle $r_i^{(\epsilon)},$ of dimension 
$$\geq (w-2)\cdot (h-2) \geq \frac{k-2}{k}w \cdot \frac{k-2}{k}h = \left( \frac{k-2}{k} \right)^d |r_i|.$$

%Hence, we have $\gamma > k\epsilon > 4\epsilon$.
%Suppose to consider for the sake of simplicity one $r_i \in \mathcal{R}$ (depicted in blue). The proposition states that $\forall r_i \in \mathcal{R}$, we have that $r_i$ is $\gamma$-bounded, i.e., each interval that composes $r_i$ is at least of size $\gamma$ and, in addition, there is a solution $\mathcal{R}^\epsilon$ of size $\geq \left(\frac{k-2}{k}\right)^d ||{\cal R}||$. In particular, in the image, we report $\mathcal{R}^\epsilon$ with a green hyperrectangle, and we notice that first of all, the intervals that compose $r_i^\epsilon$ are all $\epsilon$-aligned, i.e., each side of the interval starts precisely in correspondence of the $\epsilon$ discretization. Moreover, we have that $||R^\epsilon|| = ||r_i^\epsilon|| = 2\epsilon \cdot 3\epsilon$. From the fact that $2\epsilon \geq \frac{k-2}{k}\cdot 3\epsilon$ and similarly $3\epsilon \geq \frac{k-2}{k}\cdot 4\epsilon$ we have that $||R^\epsilon|| = ||r_i^\epsilon|| \geq \left(\frac{k-2}{k}\right)^d ||{\cal R}||$.}

The following theorem summarizes the coverage approximation guarantee and the confidence guarantee on the safety nature of the areas returned by \texttt{$\epsilon$-ProVe}.

\begin{theorem}\label{th_eProVe}
Fix a positive integer $d$ and real values $\epsilon, \alpha, R \in (0,1),$ with $R > 1-\epsilon^d.$ Let ${\cal T}$ be an instance of  the \textit{AllDNN-Verification} with input in\footnote{This assumption is w.l.o.g. modulo some normalization.} $[0,1]^d,$ and  let $k$ be the largest integer such that
$\Gamma({\cal T})$ can be partitioned into $k\epsilon$-bounded rectilinear hyperrectangle. 
%Let ${\cal R}^{(\epsilon)} = \{r_i^{(\epsilon)} \mid i=1, \dots, m\}$ be the set of areas returned by \texttt{$\epsilon$-ProVe} using $n$ samples at each iteration, with  
%$n \geq \log_R(1-\alpha^{1/m}).$

Let ${\cal R}^{(\epsilon)}_{+}$ and 
${\cal R}^{(\epsilon)}_{-}$ 
%= \%{r_i^{(\epsilon)} \mid i=1, \dots, m\}$ 
be the sets of areas identified as safe and unsafe, respectively, by \texttt{$\epsilon$-ProVe} using $n$ samples at each iteration, with  
$n \geq \log_R(1-\alpha^{1/m})$ and 
$m \geq \max\{|{\cal R}^{(\epsilon)}_{+}|, 
|{\cal R}^{(\epsilon)}_{-}|\}.$
 Then, with probability $\geq \alpha$
 \begin{enumerate}
\item (coverage guarantee) 
%with probability {\color{blue}$(1-R^n)^m$}
the solution ${\cal R}^{(\epsilon)}_{+}$ is a  $\left(\frac{k-2}{k}\right)^d$ approximation of $\Gamma({\cal T}),$ i.e., 
 $||{\cal R}^{(\epsilon)}_{+}|| \geq \left(\frac{k-2}{k}\right)^d |\Gamma({\cal T})|$; 
 \item (safety  guarantee) 
 %with probability $\geq \alpha$, 
 in each
 hyperrectangle $r \in {\cal R}^{(\epsilon)}_{+}$ at most $(1-R)\cdot|r|$ points are not safe.
 \end{enumerate}
\end{theorem}

This theorem gives two types of guarantees on the solution returned by \texttt{$\epsilon$-ProVe}. Specifically, point 2. states that for any $R < 1$ and $\alpha < 1$, 
\texttt{$\epsilon$-ProVe} can guarantee that with probability $\alpha$ no more than $(1-R)$ of the points classified as safe can, in fact, be violations. Moreover, point 1. 
guarantees that, provided the space of safety points is not too scattered---formalized by the existence of some representation in $k\epsilon$-bounded hyperrectangles--- the total area returned by $\epsilon$-ProVe is guaranteed to be close to the actual $\Gamma({\cal T}).$ 

Finally, the theorem shows that the two guarantees are attainable in an efficient way, providing a quantification of the size $n$ of the sample needed at each iteration. Note that the value of $m$ needed in defining $n$  can be either set using the upper limit $2^{d\log(1/\epsilon)}$--- which is the maximum number of possible split operations performed before reaching the $\epsilon$-precision limit---or $m$ can be
estimated by a standard doubling technique:  repeatedly  run the algorithm doubling the estimate for $m$ at each new run until the actual number of areas returned is upper bounded by the current guess for $m$. 

%\par
\noindent
\begin{proof}
The safety guarantee (item 2.) is a direct consequence of Lemma \ref{lemma_Wilks}. In fact, a hyperrectangle $r$ is returned as safe if all the $n$ sampled points from $r$ 
are not violations, i.e., their output is $\geq 0$ (see (\ref{safe-unsafe-cases})).
By Lemma \ref{lemma_Wilks}, at most $(1-R)$ of the points in $r$ can give an output $<0$, with probability $\hat{\alpha} = (1-R^n)$.
%i.e., can be a violation,  where,  from the solution of (\ref{eq:Wilks}), 
%\begin{equation}\label{int-sol}\hat{\alpha} = (1-R^n).\end{equation} 
Since samples are chosen independently in different hyperrectangles, this bound on the number of violations in a hyperrectangle of $\cal R^{(\epsilon)}$ holds simultaneously for all of them with probability $\geq {\hat{\alpha}}^m.$ With 
$n \geq \log_R(1-\alpha^{1/m})$ we have $\alpha \leq {\hat{\alpha}}^m,$ i.e., the safety guarantee holds with probability $\geq \alpha$.

For the coverage guarantee, we start by noticing that under the hypotheses on $k$, Proposition \ref{prop2} guarantees the existence of a solution ${\cal R}^{\epsilon}_1$ 
made of $\epsilon$-bounded and $\epsilon$-aligned rectilinear hyperrectangles.
Let ${\cal R}^{\epsilon}_2$ be a solution obtained from ${\cal R}^{\epsilon}_1$ by 
partitioning each hyperrectangle into hyperrectangles of minimum possible size $\epsilon^d,$
each one $\epsilon$-aligned.
The first observation is that, being a solution 
%produced by \texttt{$\epsilon$-ProVe} 
made of 
$\epsilon$-bounded and $\epsilon$-aligned rectilinear hyperrectangles, 
${\cal R}^{\epsilon}_2$ is among the solutions possibly returned by \texttt{$\epsilon$-ProVe}.
%the algorithm.
We now observe that, with probability
$\geq \alpha,$ each hyperrectangles in ${\cal R}^{\epsilon}_2$ must be contained in 
some hyperrectangle $r$ in the solution ${\cal R}^{(\epsilon)}_{+}$ returned by \texttt{$\epsilon$-ProVe}.

First note that if in each iteration of \texttt{$\epsilon$-ProVe}, $r'$ keeps on being contained in an area where both safe and violation points are sampled, then eventually  $r'$ will become itself an area to analyze. At such a step, clearly every sample in $r'$ will be safe and $r'$ will be included in ${\cal R}^{(\epsilon)}_{+},$  as desired.
Therefore, the only possibility for $r'$ not to be contained in any $r \in {\cal R}^{(\epsilon)}_{+}$ is that at some iteration an area $A \supseteq r'$ is analyzed and 
all the $n$ points sampled in $A$ turn out to be violation points, so  $A$ (including $r'$) is classified unsafe (and added to
${\cal R}^{(\epsilon)}_{-}$)
by \texttt{$\epsilon$-ProVe}. However, by Lemma \ref{lemma_Wilks} with probability $(1-R^n)$ this can 
happen only if $\epsilon^d = |r'| < (1-R)|A|,$ which contradicts the hypotheses. Hence, with probability $(1-R^n)^m \geq \alpha$, no hyperrectangle of ${\cal R}^{(\epsilon)}_{-}$ contains $r',$ whence it must be contained in a 
hyperrectangle of ${\cal R}^{(\epsilon)}_{+},$ concluding the argument.
\end{proof}

\begin{table*}
\small
\centering
\begin{tabular}{|c||cccccc||cccc||cc|}
\hline
{\textbf{Instance}} &
  \multicolumn{6}{c||}{\textbf{\texttt{$\epsilon$-ProVe} } ($\alpha = 99.9\%$)} &
  \multicolumn{4}{c||}{\textbf{Exact count} or \textbf{\textit{MC sampling}}} &
  \multicolumn{2}{c|}{\textbf{Und-estimation (\% distance)}} \\
 &
  \multicolumn{2}{c}{\# Safe regions} &
  \multicolumn{2}{c}{Safe rate} &
  \multicolumn{2}{c||}{Time} &
  \multicolumn{2}{c}{Safe rate} &
  \multicolumn{2}{c||}{Time} &
  \multicolumn{2}{c|}{} \\[0.5ex] \hline
  & \multicolumn{2}{c}{\vspace{-2.5mm}} & \multicolumn{2}{c}{} & \multicolumn{2}{c||}{} & \multicolumn{2}{c}{} &
  \multicolumn{2}{c||}{} &
  \multicolumn{2}{c|}{}\\
Model\_2\_20          & \multicolumn{2}{c}{335} & \multicolumn{2}{c}{78.50\%} & \multicolumn{2}{c||}{0.4s} & \multicolumn{2}{c}{79.1\%} & 
\multicolumn{2}{c||}{234min} & \multicolumn{2}{c|}{0.74\%} \\
Model\_2\_56          & \multicolumn{2}{c}{251} & \multicolumn{2}{c}{43.69\%} & \multicolumn{2}{c||}{0.3s} & \multicolumn{2}{c}{44.46\%} & 
\multicolumn{2}{c||}{196min} & \multicolumn{2}{c|}{1.75\%} \\ \hline
%Model\_2\_68          & \multicolumn{2}{c}{142} & \multicolumn{2}{c}{29.37\%} & \multicolumn{2}{c||}{0.2} & \multicolumn{2}{c}{31.95\%} & 
%\multicolumn{2}{c||}{210min} &
%\multicolumn{2}{c|}{8\%} \\[0.5ex] \hline

& \multicolumn{2}{c}{\vspace{-2.5mm}} & \multicolumn{2}{c}{} & \multicolumn{2}{c||}{} & \multicolumn{2}{c}{} &  \multicolumn{2}{c||}{} & \multicolumn{2}{c|}{}\\
Model\_MN\_1        & \multicolumn{2}{c}{545} & \multicolumn{2}{c}{64.72\%} & \multicolumn{2}{c||}{60.6s} & \multicolumn{2}{c}{\textit{67.59\%}} &
\multicolumn{2}{c||}{\textit{0.6s}} &
\multicolumn{2}{c|}{4.24\%}      \\
Model\_MN\_2        & \multicolumn{2}{c}{1} & \multicolumn{2}{c}{100\%} & \multicolumn{2}{c||}{0.4s} & \multicolumn{2}{c}{\textit{100\%}} &
\multicolumn{2}{c||}{\textit{0.4s}} &
\multicolumn{2}{c|}{-}        \\[0.5ex] \hline
%Model\_MN\_3       & \multicolumn{2}{c}{340} & \multicolumn{2}{c}{53.13\%} & \multicolumn{2}{c||}{28.5s} & \multicolumn{3}{c}{\textit{55.91\%}} &
%\multicolumn{2}{c||}{\textit{0.5s}} & \multicolumn{2}{c|}{5\%}     \\[0.5ex] \hline
& \multicolumn{2}{c}{\vspace{-2.5mm}} & \multicolumn{2}{c}{} & \multicolumn{2}{c||}{} & \multicolumn{2}{c}{} & 
\multicolumn{2}{c||}{} &
\multicolumn{2}{c|}{}\\
$\phi_2$ ACAS Xu\_2.1 & \multicolumn{2}{c}{2462} & \multicolumn{2}{c}{97.47\%} & \multicolumn{2}{c||}{26.9s} & \multicolumn{2}{c}{99.25\%} &
\multicolumn{2}{c||}{\textit{0.6s}} &
\multicolumn{2}{c|}{1.81\%} \\
$\phi_2$ ACAS Xu\_3.3 & \multicolumn{2}{c}{1}   & \multicolumn{2}{c}{100\%}   & \multicolumn{2}{c||}{0.4s} & \multicolumn{2}{c}{\textit{100\%}}   &
\multicolumn{2}{c||}{\textit{0.5s}}   &
\multicolumn{2}{c|}{-}      \\[0.5ex] \hline
%$\phi_2$ ACAS Xu\_4.9 & \multicolumn{2}{c}{22}  & \multicolumn{2}{c}{99.19\%} & \multicolumn{2}{c||}{0.7s} & \multicolumn{3}{c}{\textit{99.86\%}} &
%\multicolumn{2}{c||}{\textit{0.02s}} &\multicolumn{2}{c|}{0.01\%} \\[0.5ex] \hline
\end{tabular}
\caption{Comparison of \texttt{$\epsilon$-ProVe} and Exact count or Monte Carlo (MC) sampling approach on different benchmark setups. Full results and other different experiments are reported in the supplementary material.}
\label{tab:results}

\end{table*}

\begin{figure}[h!]
    \centering
    \includegraphics[width=0.59\linewidth]{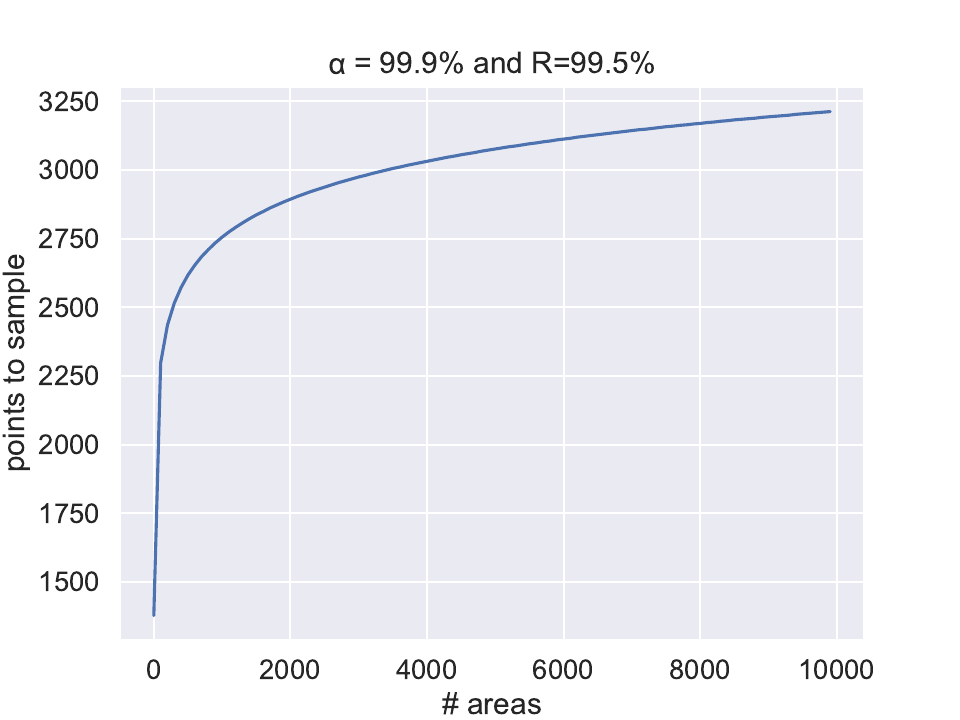}
    \includegraphics[width=0.39\linewidth]{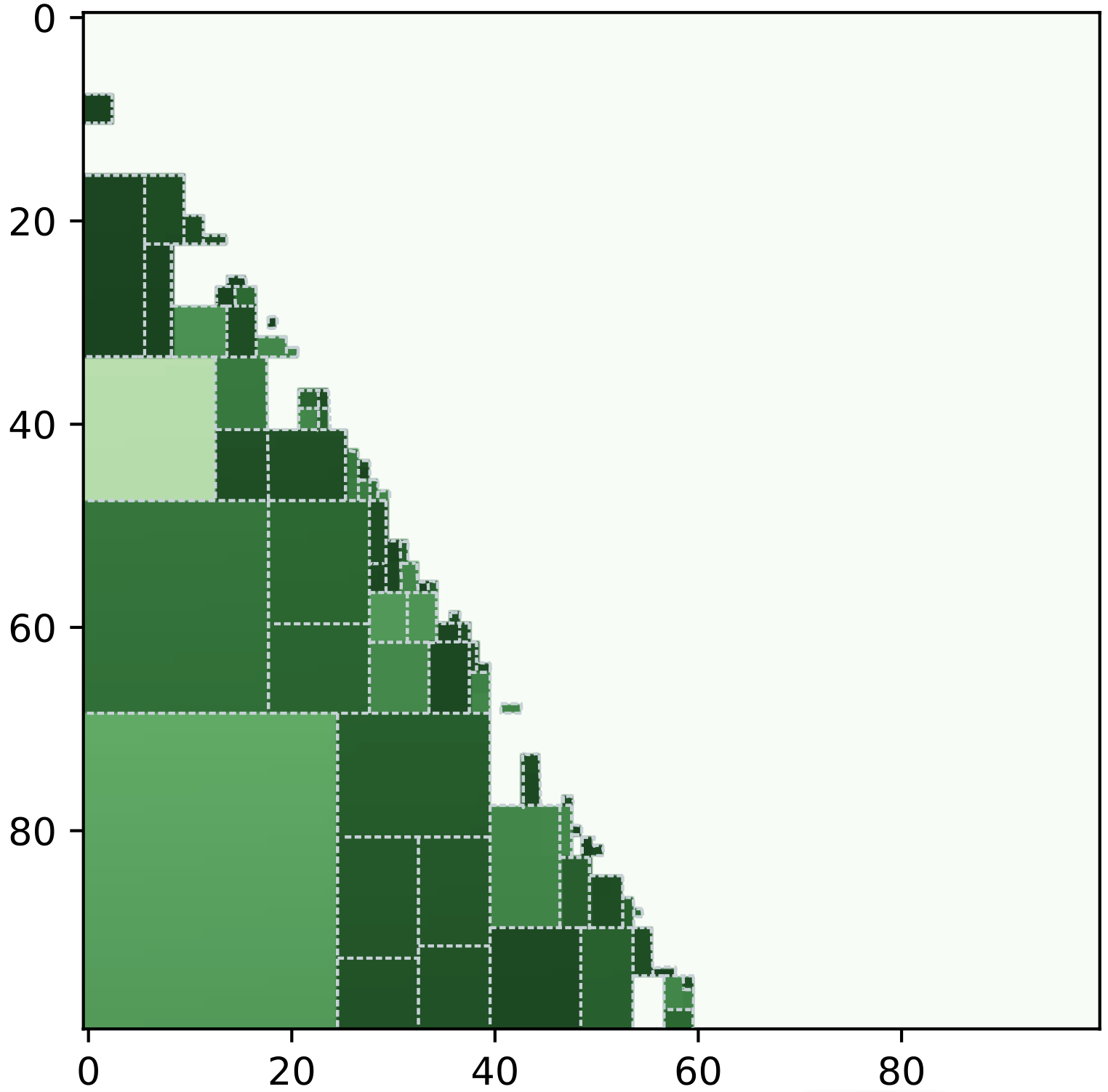}
    \caption{Left: Correlation point to sample and \# of (un)safe areas using \texttt{$\epsilon$-ProVe} to obtain a confidence $\alpha=99.9\%$ and a lower bound $R=99.5\%$. Right: example of a set of safe regions (in green) returned by \texttt{$\epsilon$-ProVe} (scaled x100).}
    \label{fig:correlation}
\end{figure}

Fig. \ref{fig:correlation} (left) shows the correlation between the number of points to be sampled based on the number of areas obtained by \texttt{$\epsilon$-ProVe} if we want to obtain a total confidence of $\alpha= 99.9\%$ and a lower bound $R=99.5\%$. As we can notice from the plot, if we compute our output reachable set sampling $n=3250$ points, we are able to obtain the desired confidence and lower bound if the number of regions is in $[1, 10000]$. For this reason, in all our empirical evaluations, we use $n=3500$ to compute $\mathcal{R}$. An 
%explanatory 
example of the possible result achievable using our approach is depicted in Fig. \ref{fig:correlation} (right) with different shades of green for better visualization.

\section{Empirical Evaluation}\label{empirical_eval}

In this section, we evaluate the scalability of our approach, and we validate the theoretical guarantees discussed in the previous section. Our analysis considers both simple DNNs to analyze in detail the theoretical guarantees and two real-world scenarios to evaluate scalability. 
The first scenario is the ACAS xu \cite{ACAS}, an airborne collision avoidance system for aircraft, which is a well-known standard benchmark for formal verification of DNNs \cite{Liu,Reluplex,Neurify}. The second scenario considers DNN trained and employed for autonomous mapless navigation tasks in a Deep Reinforcement Learning (DRL) context \cite{navigation,icra_safety, aamas_safety}. 

All the data are collected on a commercial PC equipped with an M2 Apple silicon. The code used to collect the results and several additional experiments and discussions on the impact of different heuristics for our approximation are available in the supplementary material.

\subsection{Correctness and Scalability Experiments}

These experiments aim to estimate the correctness and scalability of our approach. Specifically for each model tested, we used \texttt{$\epsilon$-ProVe} to return the set of safe regions in the domain of the property under consideration.
%\footnote{we recall that similarly can be done with the set of unsafe areas}. 
All data are collected with parameters $\alpha_{TOT} = 99.9\%$ and $R=99.5\%$ and $n=3500$ points to compute the reachable set used for the analysis. The results are presented in Table \ref{tab:results}. 
For all experiments, we report the number of safe regions returned by  \texttt{$\epsilon$-ProVe} (for which we also know the hyperrectangles position in the property domain), the percentage of safe areas relative to the total starting area (i.e., the safe rate), and the computation time. Moreover, we include a comparison, measured as percentage distance, of the safe rate computed with alternative methods, such as an exact enumeration method (whenever feasible due to the scalability issue discussed above) and a Monte Carlo (MC) Sampling approach using a large number of samples (i.e., 1 million). It's important to note that the MC sampling only provides a probabilistic estimate of the safe rate, lacking information about the location of safe regions in the input domain.

The first block of Table \ref{tab:results} involves two-dimensional models with two hidden layers of 32 nodes activated with ReLU. The safety property consists of all the intervals of $\mathcal{P}$ in the range $[0, 1]$ and a postcondition $\mathcal{Q}$ that encodes a strictly positive output. Notably, \texttt{$\epsilon$-ProVe} is able to return the set of safe regions in a fraction of a second, and the safe rate returned by our approximation deviates at most a $1.75\%$ from the one computed by an exact count, which shows the tightness of the bound returned by our approach. In the second block of Tab. \ref{tab:results}, the Mapless Navigation (MN) DNNs are composed of 22 inputs, two hidden layers of 64 nodes activated with $ReLU$, and finally, an output space composed of five nodes, which encode the possible actions of the robot. We test a behavioral safety property where $\mathcal{P}$ encodes a potentially unsafe situation (e.g., \textit{there is an obstacle in front}), and the postcondition $\mathcal{Q}$ specifies the unsafe action that should not be selected.  
%($\bigvee_{i=2,3} y_1 < y_i$ if $y_1$ encodes a forward movement). 
The table illustrates how increasing input space and complexity affects computation time. Nevertheless, the proposed approximation remains efficient even for ACAS xu tests, returning results within seconds. Crucially, focusing on \textit{Model\_MN\_2} and $\phi_2$ ACAS Xu\_3.3,
\texttt{$\epsilon$-ProVe} states that all the property's domain is safe (i.e., no violation points). The correctness of the results was verified by employing VeriNet \cite{verinet}, a state-of-the-art FV tool.
%that returns an \texttt{UNSAT} answer, meaning that both properties domains are actually provable safe, thus confirming the correctness of our result.

% \subsection{\#DNN-Verification on Safe Regions}
% To get further empirical confirmation of the correctness of our results, we perform an additional experiment on the individual areas deemed safe by our approximation. In detail, given a particular set of safe areas returned by \texttt{$\epsilon$-ProVe}, we employ on each region a state-of-the-art exact count verifier called ProVe \cite{ProVe} and show that in all of these regions $A$, the number of violation points never exceeds $(1-R)\cdot\vert A\vert$. The results obtained (reported in the supplementary material) show that all the regions are actually 100\% provably safe, confirming the correctness of the lower bound returned.
\vspace{-2mm}
\section{Discussion}

We studied the \textit{AllDNN-Verification}, a novel problem in the FV of DNNs asking for the set of all the safe regions for a given safety property.
Due to the \#P-hardness of the problem, we proposed an approximation approach, \texttt{$\epsilon$-ProVe}, 
%which constructs a solution made of rectilinear hyperrectangles. 
%\texttt{$\epsilon$-ProVe} 
which is, to the best of our knowledge, the first method able to efficiently approximate the safe regions
%of a DNN w.r.t. a given property 
with some 
%theoretical probabilistic 
guarantees on the tightness of the solution returned.
%tightness of the area returned and on  probabilistic lower bound of the safe areas for a given safety specification. 
%We first mathematically proved the correctness and the tightness of our solution, and finally, we empirically analyzed the correctness and scalability of our approach on a set of benchmarks, including a real-world mapless navigation problem and the ACAS Xu benchmark, considered a standard in the formal verification community. The proposed method balances computational efficiency and reliable results, offering valuable insights into the provable (un)safe regions of a DNN's property input domain. 
We believe \texttt{$\epsilon$-ProVe} is an important step to provide consistent and effective tools for analyzing safety in DNNs. 

An interesting future direction could be testing the enumeration of the unsafe regions for subsequent patching or a safe retrain on the areas discovered by our approach. 

\clearpage
\bibliography{aaai24}

\clearpage
\appendix
\section{Supplementary Material}

\subsection{Impact of Heuristics on \texttt{$\epsilon$-ProVe} Performances}

Although the correctness of the approximation does not depend on the heuristic chosen to split the input space under consideration (as shown in Theorem \ref{th_eProVe}), we performed an additional experiment to understand its impact on the quality of the estimation. In particular, as shown in Alg. \ref{alg:eProVe}, the loop ends when either \texttt{$\epsilon$-ProVe} explores all the unknown regions, or it reaches a specific $\epsilon-precision$. Regarding the latter, all our results are given for the case of a maximum number of splits equal to $s=18$, since, in preliminary testing, this value has shown to give the best trade-off between efficiency and accuracy. 
Hence, with respect to theoretical analysis in Theorem \ref{th_eProVe}), with this choice, we expect the solution returned by $\epsilon$-ProVe to be 
 $\epsilon$-bounded, for some $\epsilon \in [2^{1/s}, 2^{d/s}]$ considering the maximum and the expected number of times that a split happens on the same dimension.

Moreover, another key aspect when using iterative refinement is the choice of the dimension (input coordinate) and the position where to perform the split. To this end, our sampling-based approach to estimating the reachable set also allows us to obtain some useful statistical measures, such as the mean and median of the samples in each dimension, that can be useful for the heuristics. More specifically, we tested the following five heuristics:

\begin{figure}[b]
        \centering
        \includegraphics[width=0.8\linewidth]{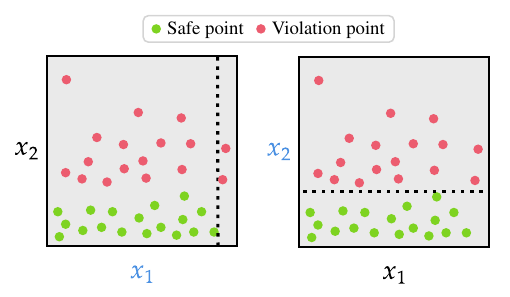}
        \vspace{-4mm}
        \caption{Example of single test split on each dimension $x_1,x_2,$ (highlighted in blue) using $H5$ in a specific situation.}
        \label{fig:h5}
    \end{figure}
\begin{itemize}
    \item \textit{H1}: split the node along the bigger dimension, i.e., $d = argmax_{i=1}^m u_i-\ell_i$, with $u_i$ and $\ell_i$ the upper and lower bound of each interval, respectively. The split is performed in correspondence with the median value of the samples that result safe after the propagation through the DNN.
    \item \textit{H2}: as \textit{H1}, but the split is performed along the mean value of the safe samples.
    \item \textit{H3}: select one dimension randomly, and the split is performed along the median value of the safe samples.
    \item \textit{H4}: as \textit{H3}, but the split is performed along the mean value of the safe samples.
    \item \textit{H5}: select the dimension and the position to split based on the distribution of the safe and violation points, i.e., partitioning the input space into parts where we have as many as possible violation points in one part and safety in the other. We report a simplified example of this heuristic in Fig. \ref{fig:h5}.

    More specifically, let us suppose to have only two dimensions and to have sampled $n$ points that result in the situation depicted in Fig. \ref{fig:h5}. Hence, for each dimension, the heuristic seeks to find the splitting point that better divides the safe points from the unsafe (violation) ones. In this situation, the dimension chosen will be $x_2$ as it allows a better partition of the input space, splitting in correspondence with the dotted line, which represents the maximum value of the safe points along $x_2$.
    \end{itemize}

    \begin{table}[t]
\centering
\begin{tabular}{ccccccc}
\hline
\multicolumn{1}{|c||}{\multirow{2}{*}{\textbf{Instance}}} &

  \multicolumn{3}{c|}{\textbf{H1}}\\ 
\multicolumn{1}{|c||}{} &
  \multicolumn{1}{c}{\# Safe regions} &
  \multicolumn{1}{c}{Safe rate} &
  \multicolumn{1}{c|}{Time} \\ \hline
  \multicolumn{1}{|c||}{\vspace{-2mm}} &
  \multicolumn{1}{c}{} &
  \multicolumn{1}{c}{} &
  \multicolumn{1}{c|}{} \\
\multicolumn{1}{|c||}{Model\_2\_20} &
  \multicolumn{1}{c}{731} &
  \multicolumn{1}{c}{78.91\%} &
  \multicolumn{1}{c|}{0.5s} \\ 
\multicolumn{1}{|c||}{Model\_MN\_3} &
  \multicolumn{1}{c}{1196} &
  \multicolumn{1}{c}{53.14\%} &
\multicolumn{1}{c|}{1m21s} \\ 
\multicolumn{1}{|c||}{$\phi_2$ ACAS Xu\_2.1} &
  \multicolumn{1}{c}{3540} &
  \multicolumn{1}{c}{97.32\%} &
  \multicolumn{1}{c|}{29.1s}\\[0.5ex] \hline
  \multicolumn{1}{c||}{} &
  \multicolumn{3}{c|}{\textbf{H2}}\\ 
\multicolumn{1}{c||}{} &
  \multicolumn{1}{c}{\# Safe regions} &
  \multicolumn{1}{c}{Safe rate} &
  \multicolumn{1}{c|}{Time} \\ \hline
  \multicolumn{1}{|c||}{\vspace{-2mm}} &
  \multicolumn{1}{c}{} &
  \multicolumn{1}{c}{} &
  \multicolumn{1}{c|}{} \\
\multicolumn{1}{|c||}{Model\_2\_20} &
  \multicolumn{1}{c}{678} &
  \multicolumn{1}{c}{79.01\%} &
  \multicolumn{1}{c|}{1.2s} \\ 
\multicolumn{1}{|c||}{Model\_MN\_3} &
  \multicolumn{1}{c}{739} &
  \multicolumn{1}{c}{53.08\%} &
\multicolumn{1}{c|}{1m42s} \\ 
\multicolumn{1}{|c||}{$\phi_2$ ACAS Xu\_2.1} &
  \multicolumn{1}{c}{1613} &
  \multicolumn{1}{c}{97.24\%} &
  \multicolumn{1}{c|}{39s}\\[0.5ex] \hline
  
\multicolumn{1}{c||}{} &
  \multicolumn{3}{c|}{\textbf{H3}} \\ 
\multicolumn{1}{c||}{} &
  \multicolumn{1}{|c}{\# Safe regions} &
  \multicolumn{1}{c}{Safe rate} &
  \multicolumn{1}{c|}{Time}  \\ \hline
  \multicolumn{1}{|c||}{\vspace{-2mm}} &
  \multicolumn{1}{c}{} &
  \multicolumn{1}{c}{} &
  \multicolumn{1}{c|}{} \\
\multicolumn{1}{|c||}{Model\_2\_20} &
  \multicolumn{1}{c}{2319} &
  \multicolumn{1}{c}{78.63\%} &
  \multicolumn{1}{c|}{1.9s} \\ 
\multicolumn{1}{|c||}{Model\_MN\_3} &
  \multicolumn{1}{c}{-} &
  \multicolumn{1}{c}{-} &
  \multicolumn{1}{c|}{-}\\ 
\multicolumn{1}{|c||}{$\phi_2$ ACAS Xu\_2.1} &
  \multicolumn{1}{c}{22853} &
  \multicolumn{1}{c}{83\%} &
  \multicolumn{1}{c|}{4m20s}\\[0.5ex] \hline
 \multicolumn{1}{c||}{} &
  \multicolumn{3}{c|}{\textbf{H4}} \\ 
\multicolumn{1}{c||}{} &
  \multicolumn{1}{|c}{\# Safe regions} &
  \multicolumn{1}{c}{Safe rate} &
  \multicolumn{1}{c|}{Time}  \\ \hline
  \multicolumn{1}{|c||}{\vspace{-2mm}} &
  \multicolumn{1}{c}{} &
  \multicolumn{1}{c}{} &
  \multicolumn{1}{c|}{} \\
\multicolumn{1}{|c||}{Model\_2\_20} &
  \multicolumn{1}{c}{1757} &
  \multicolumn{1}{c}{78.8\%} &
  \multicolumn{1}{c|}{2.9s} \\ 
\multicolumn{1}{|c||}{Model\_MN\_3} &
  \multicolumn{1}{c}{-} &
  \multicolumn{1}{c}{-} &
  \multicolumn{1}{c|}{-}\\ 
\multicolumn{1}{|c||}{$\phi_2$ ACAS Xu\_2.1} &
  \multicolumn{1}{c}{31515} &
  \multicolumn{1}{c}{78.09\%} &
  \multicolumn{1}{c|}{5m}\\[0.5ex] \hline
  \multicolumn{1}{c||}{} &
  \multicolumn{3}{c|}{\textbf{H5}} \\ 
\multicolumn{1}{c||}{} &
  \multicolumn{1}{|c}{\# Safe regions} &
  \multicolumn{1}{c}{Safe rate} &
  \multicolumn{1}{c|}{Time}  \\ \hline
  \multicolumn{1}{|c||}{\vspace{-2mm}} &
  \multicolumn{1}{c}{} &
  \multicolumn{1}{c}{} &
  \multicolumn{1}{c|}{} \\
\multicolumn{1}{|c||}{Model\_2\_20} &
  \multicolumn{1}{c}{\textbf{355}} &
  \multicolumn{1}{c}{\textbf{78.5\%}} &
  \multicolumn{1}{c|}{\textbf{0.4s}} \\ 
\multicolumn{1}{|c||}{Model\_MN\_3} &
  \multicolumn{1}{c}{\textbf{845}} &
  \multicolumn{1}{c}{\textbf{52.95\%}} &
  \multicolumn{1}{c|}{\textbf{1m15s}}\\ 
\multicolumn{1}{|c||}{$\phi_2$ ACAS Xu\_2.1} &
  \multicolumn{1}{c}{\textbf{2462}} &
  \multicolumn{1}{c}{\textbf{97.47\%}} &
  \multicolumn{1}{c|}{\textbf{26.9s}}\\[0.5ex] \hline

\end{tabular}
\caption{Comparison of different heuristic on \texttt{$\epsilon$-Prove}. '-' indicates a heuristic that requires more than 5 minutes. The target safe rates obtained using either an exact enumeration method (when possible) or a Monte Carlo estimation are: for model\_2\_20 79.1\%, for model\_MN\_3 55.93\% and finally for $\phi_2$ ACAS Xu\_2.1 99.25\%. }
\label{tab:ablation_heu}
\vspace{-3mm}
\end{table}

\begin{table*}

\centering
\begin{tabular}{ccccccc}
\hline
\multicolumn{1}{|c||}{\multirow{2}{*}{\textbf{Hyperrectangle $r$}}} &

  \multicolumn{3}{c|}{\textbf{ProVe}}\\ 
\multicolumn{1}{|c||}{} &
  \multicolumn{3}{c|}{Violation rate}\\ \hline
  \multicolumn{1}{|c||}{\vspace{-2mm}} &
  \multicolumn{1}{c}{} &
  \multicolumn{1}{c}{} &
  \multicolumn{1}{c|}{} \\
\multicolumn{1}{|c||}
{[[0.6999231, 1],	[0, 0.2563121]]
} &
  \multicolumn{3}{c|}{0\%}\\[0.5ex] 
  \multicolumn{1}{|c||}{[[0.6999231, 1],[0.2563121, 0.40074086]]} &
  \multicolumn{3}{c|}{0\%}\\[0.5ex] 
  \multicolumn{1}{|c||}{[[0.48361894, 0.6999231],	[0, 0.18614466]]
} &
  \multicolumn{3}{c|}{0\%}\\[0.5ex] 
  \multicolumn{1}{|c||}{[[0.23474114, 0.34573981],[0, 0.09308449]]
} &
  \multicolumn{3}{c|}{0\%}\\[0.5ex]

  \multicolumn{1}{|c||}
  {[[0.39214072, 0.39503244],[0.245827,  0.25156769]]} &
  \multicolumn{3}{c|}{\textbf{0.031\%}}\\[0.5ex]

  \multicolumn{1}{|c||}
  {[[0.95905697, 0.95924747],	[0.60187447, 0.60229677]]} &
  \multicolumn{3}{c|}{\textbf{0.0488\%}}\\[0.5ex]

  \hline

\end{tabular}
\caption{Partial results of Prove \cite{ProVe} on each single hyperrectangle $r$ returned as solution for the \textit{$\epsilon$-RUA-DNN} for $model\_2\_68$. The csv file with full results is available in the supplementary material folder. }
\label{tab:exact_eval}
\end{table*}

We have noticed that all these heuristics have roughly the same computational cost on a single split, and thus an improvement in the computation time of the complete execution of \texttt{$\epsilon$-ProVe} translates directly into the choice of better heuristics. We tested each heuristic on three different models with $\alpha=99.9\%, R=99.5\%$, and $n=3500$ samples. For each heuristic tested, we collected: the number of safe regions, the safe rate, and the computation time, respectively. Ideally, for a similar safe rate, the best heuristic should return the fewest regions (i.e., the best compact solution of the safe regions) in the lowest possible time. In Tab.\ref{tab:ablation_heu}, we report the results obtained. As we can see, \textit{H5} is the best heuristic as it yields the best trade-off between the efficiency and compactness of the solution for all the models tested. We can notice that for $model\_MN\_3$ the improvement between $H2$ and $H5$ is not significant, we think this is due to the larger cardinality of the input space w.r.t the other models. Further experiments should be performed to devise better heuristics for these types of DNNs.

\subsection{Full Results \texttt{$\epsilon$-ProVe} on Different Benchmarks}

We report in Tab\;\ref{tab:fullresults} the full results comparing \texttt{$\epsilon$-ProVe} on different benchmarks. As stated in the main paper, we consider three different setups: the first one is a set of small DNNs, the second one is a realistic set of Mapless Navigation (MN) DNNs, used in a DRL context, and finally, we used the standard FV benchmark ACAS xu\cite{ACAS}. For the latter, we consider only a subset of the original 45 models. More precisely, based on the result of the second international Verification of Neural Networks (VNN) competition \cite{VNNcomp}, we test part of the models for which the property $\phi_2$ does not hold (i.e., the models that present at least one single input configuration that violates the safety property) and one single model ($\phi_2$ ACAS Xu\_3.3) for which the property holds (i.e., 100\% provable safe). 

\begin{table*}
\centering
\begin{tabular}{|c||cccccc||cccc||cc|}
\hline
{\textbf{Instance}} &
  \multicolumn{6}{c||}{\textbf{\texttt{$\epsilon$-ProVe} } ($\alpha_{TOT} = 99.9\%$)} &
  \multicolumn{4}{c||}{\textbf{Exact count} or \textbf{\textit{MC sampling}}} &
  \multicolumn{2}{c|}{\textbf{Und-estimation (\% distance)}} \\
 &
  \multicolumn{2}{c}{\# Safe regions} &
  \multicolumn{2}{c}{Safe rate} &
  \multicolumn{2}{c||}{Time} &
  \multicolumn{2}{c}{Safe rate} &
  \multicolumn{2}{c||}{Time} &
  \multicolumn{2}{c|}{} \\[0.5ex] \hline
  & \multicolumn{2}{c}{\vspace{-2.5mm}} & \multicolumn{2}{c}{} & \multicolumn{2}{c||}{} & \multicolumn{2}{c}{} &
  \multicolumn{2}{c||}{} &
  \multicolumn{2}{c|}{}\\
Model\_2\_20          & \multicolumn{2}{c}{335} & \multicolumn{2}{c}{78.50\%} & \multicolumn{2}{c||}{0.4s} & \multicolumn{2}{c}{79.1\%} & 
\multicolumn{2}{c||}{234min} & \multicolumn{2}{c|}{0.74\%} \\
Model\_2\_56          & \multicolumn{2}{c}{251} & \multicolumn{2}{c}{43.69\%} & \multicolumn{2}{c||}{0.3s} & \multicolumn{2}{c}{44.46\%} & 
\multicolumn{2}{c||}{196min} & \multicolumn{2}{c|}{1.75\%} \\
Model\_2\_68          & \multicolumn{2}{c}{252} & \multicolumn{2}{c}{31.07\%} & \multicolumn{2}{c||}{0.3s} & \multicolumn{2}{c}{31.72\%} & 
\multicolumn{2}{c||}{210min} &
\multicolumn{2}{c|}{2.07\%} \\ \hline

& \multicolumn{2}{c}{\vspace{-2.5mm}} & \multicolumn{2}{c}{} & \multicolumn{2}{c||}{} & \multicolumn{2}{c}{} &  \multicolumn{2}{c||}{} & \multicolumn{2}{c|}{}\\
Model\_MN\_1        & \multicolumn{2}{c}{545} & \multicolumn{2}{c}{64.72\%} & \multicolumn{2}{c||}{60.6s} & \multicolumn{2}{c}{\textit{67.59\%}} &
\multicolumn{2}{c||}{\textit{0.6s}} &
\multicolumn{2}{c|}{4.24\%}       \\
Model\_MN\_2        & \multicolumn{2}{c}{1} & \multicolumn{2}{c}{100\%} & \multicolumn{2}{c||}{0.4s} & \multicolumn{2}{c}{\textit{100\%}} &
\multicolumn{2}{c||}{\textit{0.4s}} &
\multicolumn{2}{c|}{-}       \\
Model\_MN\_3       & \multicolumn{2}{c}{845} & \multicolumn{2}{c}{52.95\%} & \multicolumn{2}{c||}{75.5s} & \multicolumn{2}{c}{\textit{55.93\%}} &
\multicolumn{2}{c||}{\textit{0.4s}} & \multicolumn{2}{c|}{5.33\%} \\ \hline
& \multicolumn{2}{c}{\vspace{-2.5mm}} & \multicolumn{2}{c}{} & \multicolumn{2}{c||}{} & \multicolumn{2}{c}{} & 
\multicolumn{2}{c||}{} &
\multicolumn{2}{c|}{}\\
$\phi_2$ ACAS Xu\_2.1 & \multicolumn{2}{c}{2462} & \multicolumn{2}{c}{97.47\%} & \multicolumn{2}{c||}{26.9s} & \multicolumn{2}{c}{99.25\%} &
\multicolumn{2}{c||}{\textit{0.6s}} &
\multicolumn{2}{c|}{1.81\%} \\
$\phi_2$ ACAS Xu\_2.2 & \multicolumn{2}{c}{1990}   & \multicolumn{2}{c}{97.12\%}   & \multicolumn{2}{c||}{21.7s} & \multicolumn{2}{c}{\textit{98.66\%}}   &
\multicolumn{2}{c||}{\textit{0.5s}}   &
\multicolumn{2}{c|}{1.56\%}      
\\ 

$\phi_2$ ACAS Xu\_2.3 & \multicolumn{2}{c}{2059}   & \multicolumn{2}{c}{96.77\%}   & \multicolumn{2}{c||}{20.0s} & \multicolumn{2}{c}{\textit{98.22\%}}   &
\multicolumn{2}{c||}{\textit{0.5s}}   &
\multicolumn{2}{c|}{1.47\%}      
\\ 

$\phi_2$ ACAS Xu\_2.4 & \multicolumn{2}{c}{2451}   & \multicolumn{2}{c}{97.97\%}   & \multicolumn{2}{c||}{17.5s} & \multicolumn{2}{c}{\textit{99.09\%}}   &
\multicolumn{2}{c||}{\textit{0.5s}}   &
\multicolumn{2}{c|}{1.13\%}      
\\ 

$\phi_2$ ACAS Xu\_2.5 & \multicolumn{2}{c}{2227}   & \multicolumn{2}{c}{95.72\%}   & \multicolumn{2}{c||}{23.6s} & \multicolumn{2}{c}{\textit{98.19\%}}   &
\multicolumn{2}{c||}{\textit{0.5s}}   &
\multicolumn{2}{c|}{2.51\%}      
\\ 
$\phi_2$ ACAS Xu\_2.6 & \multicolumn{2}{c}{1745}   & \multicolumn{2}{c}{97.46\%}   & \multicolumn{2}{c||}{19.2s} & \multicolumn{2}{c}{\textit{98.79\%}}   &
\multicolumn{2}{c||}{\textit{0.5s}}   &
\multicolumn{2}{c|}{1.34\%}      
\\ 
$\phi_2$ ACAS Xu\_2.7 & \multicolumn{2}{c}{2165}   & \multicolumn{2}{c}{95.69\%}   & \multicolumn{2}{c||}{23.6s} & \multicolumn{2}{c}{\textit{97.35\%}}   &
\multicolumn{2}{c||}{\textit{0.5s}}   &
\multicolumn{2}{c|}{1.7\%}      
\\ 
$\phi_2$ ACAS Xu\_2.8 & \multicolumn{2}{c}{1101}   & \multicolumn{2}{c}{97.52\%}   & \multicolumn{2}{c||}{9.0s} & \multicolumn{2}{c}{\textit{98.06\%}}   &
\multicolumn{2}{c||}{\textit{0.5s}}   &
\multicolumn{2}{c|}{0.55\%}      
\\ 
$\phi_2$ ACAS Xu\_2.9 & \multicolumn{2}{c}{767}   & \multicolumn{2}{c}{99.24\%}   & \multicolumn{2}{c||}{6.7s} & \multicolumn{2}{c}{\textit{99.7\%}}   &
\multicolumn{2}{c||}{\textit{0.5s}}   &
\multicolumn{2}{c|}{0.47\%}      
\\ 
$\phi_2$ ACAS Xu\_3.1 & \multicolumn{2}{c}{1378}   & \multicolumn{2}{c}{97.37\%}   & \multicolumn{2}{c||}{13.1s} & \multicolumn{2}{c}{\textit{98.15\%}}   &
\multicolumn{2}{c||}{\textit{0.5s}}   &
\multicolumn{2}{c|}{0.79\%}      
\\ 
$\phi_2$ ACAS Xu\_3.3 & \multicolumn{2}{c}{1}   & \multicolumn{2}{c}{100\%}   & \multicolumn{2}{c||}{0.4s} & \multicolumn{2}{c}{\textit{100\%}}   &
\multicolumn{2}{c||}{\textit{0.5s}}   &
\multicolumn{2}{c|}{-}      
\\ 
$\phi_2$ ACAS Xu\_3.4 & \multicolumn{2}{c}{1230}   & \multicolumn{2}{c}{99.03\%}   & \multicolumn{2}{c||}{9.3s} & \multicolumn{2}{c}{\textit{99.53\%}}   &
\multicolumn{2}{c||}{\textit{0.5s}}   &
\multicolumn{2}{c|}{0.49\%}      
\\ 
$\phi_2$ ACAS Xu\_3.5 & \multicolumn{2}{c}{1166}   & \multicolumn{2}{c}{98.39\%}   & \multicolumn{2}{c||}{8.0s} & \multicolumn{2}{c}{\textit{98.9\%}}   &
\multicolumn{2}{c||}{\textit{0.5s}}   &
\multicolumn{2}{c|}{0.51\%}      
\\

$\phi_2$ ACAS Xu\_3.6 & \multicolumn{2}{c}{1823}   & \multicolumn{2}{c}{96.34\%}   & \multicolumn{2}{c||}{20.8s} & \multicolumn{2}{c}{\textit{98.17\%}}   &
\multicolumn{2}{c||}{\textit{0.4s}}   &
\multicolumn{2}{c|}{1.86\%}      
\\ 
$\phi_2$ ACAS Xu\_3.7 & \multicolumn{2}{c}{1079}   & \multicolumn{2}{c}{98.16\%}   & \multicolumn{2}{c||}{11.1s} & \multicolumn{2}{c}{\textit{99.82\%}}   &
\multicolumn{2}{c||}{\textit{0.5s}}   &
\multicolumn{2}{c|}{0.66\%}      
\\ 
$\phi_2$ ACAS Xu\_3.8 & \multicolumn{2}{c}{1854}   & \multicolumn{2}{c}{97.97\%}   & \multicolumn{2}{c||}{17.0s} & \multicolumn{2}{c}{\textit{99.09\%}}   &
\multicolumn{2}{c||}{\textit{0.5s}}   &
\multicolumn{2}{c|}{1.12\%}      
\\ 
$\phi_2$ ACAS Xu\_3.9 & \multicolumn{2}{c}{1374}   & \multicolumn{2}{c}{95.96\%}   & \multicolumn{2}{c||}{17.6s} & \multicolumn{2}{c}{\textit{97.34\%}}   &
\multicolumn{2}{c||}{\textit{0.5s}}   &
\multicolumn{2}{c|}{1.43\%}      
\\ 
$\phi_2$ ACAS Xu\_4.1 & \multicolumn{2}{c}{1205}  & \multicolumn{2}{c}{98.91\%} & \multicolumn{2}{c||}{10.1s} & \multicolumn{2}{c}{\textit{99.6\%}} &
\multicolumn{2}{c||}{\textit{0.5s}} &\multicolumn{2}{c|}{0.67\%} \\

$\phi_2$ ACAS Xu\_4.3 & \multicolumn{2}{c}{2395}  & \multicolumn{2}{c}{97.38\%} & \multicolumn{2}{c||}{18.9s} & \multicolumn{2}{c}{\textit{98.56\%}} &
\multicolumn{2}{c||}{\textit{0.5s}} &\multicolumn{2}{c|}{1.2\%}\\

$\phi_2$ ACAS Xu\_4.4 & \multicolumn{2}{c}{2693}  & \multicolumn{2}{c}{97.99\%} & \multicolumn{2}{c||}{17.7s} & \multicolumn{2}{c}{\textit{99\%}} &
\multicolumn{2}{c||}{\textit{0.5s}} &\multicolumn{2}{c|}{1.04\%}\\

$\phi_2$ ACAS Xu\_4.5 & \multicolumn{2}{c}{1733}  & \multicolumn{2}{c}{97.24\%} & \multicolumn{2}{c||}{16.2s} & \multicolumn{2}{c}{\textit{98.2\%}} &
\multicolumn{2}{c||}{\textit{0.5s}} &\multicolumn{2}{c|}{1.01\%}\\

$\phi_2$ ACAS Xu\_4.6 & \multicolumn{2}{c}{1843}  & \multicolumn{2}{c}{96.72\%} & \multicolumn{2}{c||}{17.0s} & \multicolumn{2}{c}{\textit{97.74\%}} &
\multicolumn{2}{c||}{\textit{0.5s}} &\multicolumn{2}{c|}{1.04\%}\\

$\phi_2$ ACAS Xu\_4.7 & \multicolumn{2}{c}{1923}  & \multicolumn{2}{c}{96.34\%} & \multicolumn{2}{c||}{25.7s} & \multicolumn{2}{c}{\textit{98.17\%}} &
\multicolumn{2}{c||}{\textit{0.5s}} &\multicolumn{2}{c|}{1.86\%}\\

$\phi_2$ ACAS Xu\_4.8 & \multicolumn{2}{c}{1996}  & \multicolumn{2}{c}{95.53\%} & \multicolumn{2}{c||}{22.3s} & \multicolumn{2}{c}{\textit{98.14\%}} &
\multicolumn{2}{c||}{\textit{0.5s}} &\multicolumn{2}{c|}{1.64\%}\\

$\phi_2$ ACAS Xu\_4.9 & \multicolumn{2}{c}{601}  & \multicolumn{2}{c}{99.62\%} & \multicolumn{2}{c||}{4.3s} & \multicolumn{2}{c}{\textit{99.85\%}} &
\multicolumn{2}{c||}{\textit{0.5s}} &\multicolumn{2}{c|}{0.23\%}\\

$\phi_2$ ACAS Xu\_5.1 & \multicolumn{2}{c}{2530}  & \multicolumn{2}{c}{97.77\%} & \multicolumn{2}{c||}{16.5s} & \multicolumn{2}{c}{\textit{98.72\%}} &
\multicolumn{2}{c||}{\textit{0.5s}} &\multicolumn{2}{c|}{0.97\%}\\

$\phi_2$ ACAS Xu\_5.2 & \multicolumn{2}{c}{2496}  & \multicolumn{2}{c}{96.59\%} & \multicolumn{2}{c||}{27.7s} & \multicolumn{2}{c}{\textit{98.92\%}} &
\multicolumn{2}{c||}{\textit{0.5s}} &\multicolumn{2}{c|}{2.36\%}\\

$\phi_2$ ACAS Xu\_5.4 & \multicolumn{2}{c}{2875}  & \multicolumn{2}{c}{97.83\%} & \multicolumn{2}{c||}{21.6s} & \multicolumn{2}{c}{\textit{99.13\%}} &
\multicolumn{2}{c||}{\textit{0.5s}} &\multicolumn{2}{c|}{1.31\%}\\

$\phi_2$ ACAS Xu\_5.5 & \multicolumn{2}{c}{1660}  & \multicolumn{2}{c}{97.07\%} & \multicolumn{2}{c||}{15.1s} & \multicolumn{2}{c}{\textit{98.03\%}} &
\multicolumn{2}{c||}{\textit{0.5s}} &\multicolumn{2}{c|}{0.97\%}\\

$\phi_2$ ACAS Xu\_5.6 & \multicolumn{2}{c}{1909}  & \multicolumn{2}{c}{97.06\%} & \multicolumn{2}{c||}{14.7s} & \multicolumn{2}{c}{\textit{97.94\%}} &
\multicolumn{2}{c||}{\textit{0.5s}} &\multicolumn{2}{c|}{0.89\%}\\

$\phi_2$ ACAS Xu\_5.7 & \multicolumn{2}{c}{1452}  & \multicolumn{2}{c}{96.15\%} & \multicolumn{2}{c||}{16.2s} & \multicolumn{2}{c}{\textit{97.2\%}} &
\multicolumn{2}{c||}{\textit{0.5s}} &\multicolumn{2}{c|}{1.09\%}\\

$\phi_2$ ACAS Xu\_5.8 & \multicolumn{2}{c}{2357}  & \multicolumn{2}{c}{95.15\%} & \multicolumn{2}{c||}{27.5s} & \multicolumn{2}{c}{\textit{97.74\%}} &
\multicolumn{2}{c||}{\textit{0.5s}} &\multicolumn{2}{c|}{2.65\%}\\

$\phi_2$ ACAS Xu\_5.9 & \multicolumn{2}{c}{1494}  & \multicolumn{2}{c}{97.13\%} & \multicolumn{2}{c||}{10.9s} & \multicolumn{2}{c}{\textit{97.83\%}} &
\multicolumn{2}{c||}{\textit{0.5s}} &\multicolumn{2}{c|}{0.71\%}
\\[0.5ex] \hline

& \multicolumn{2}{c}{\vspace{-2.5mm}} & \multicolumn{2}{c}{} & \multicolumn{2}{c||}{} & \multicolumn{2}{c}{} &  \multicolumn{2}{c||}{} & \multicolumn{2}{c|}{}\\

$\phi_3$ ACAS Xu\_1.3 & \multicolumn{2}{c}{1}  & \multicolumn{2}{c}{100\%} & \multicolumn{2}{c||}{0.7s} & \multicolumn{2}{c}{\textit{100\%}} &
\multicolumn{2}{c||}{\textit{0.6s}} &\multicolumn{2}{c|}{-}\\

$\phi_3$ ACAS Xu\_1.4 & \multicolumn{2}{c}{1}  & \multicolumn{2}{c}{100\%} & \multicolumn{2}{c||}{0.7s} & \multicolumn{2}{c}{\textit{100\%}} &
\multicolumn{2}{c||}{\textit{0.6s}} &\multicolumn{2}{c|}{-}\\

$\phi_3$ ACAS Xu\_1.5 & \multicolumn{2}{c}{1}  & \multicolumn{2}{c}{100\%} & \multicolumn{2}{c||}{0.7s} & \multicolumn{2}{c}{\textit{100\%}} &
\multicolumn{2}{c||}{\textit{0.6s}} &\multicolumn{2}{c|}{-}\\[0.5ex] \hline

 & \multicolumn{2}{c}{}  & \multicolumn{2}{c}{} & \multicolumn{2}{c||}{} & \multicolumn{2}{c}{\textit{}} &
\multicolumn{2}{c||}{\textit{}} &\multicolumn{2}{c|}{\textbf{Mean: 1.27\%}}
\\[0.5ex] \hline
\end{tabular}
\caption{Comparison of \texttt{$\epsilon$-ProVe} and Exact count or Monte Carlo (MC) sampling approach on different benchmark setups.}
\label{tab:fullresults}
\vspace{-3mm}
\end{table*}

Moreover, to further validate the correctness of our approximation, we also performed a final experiment on property $\phi_3$ of the same benchmark. This property is particularly interesting as it holds for most of the 45 models (as shown in \cite{VNNcomp}). In particular, in this scenario we tested only models for which the property holds, i.e., we expect a 100\% of safe rate for each model tested. For simplicity, in Tab. \ref{tab:fullresults}, we report only the first three models tested for $\phi_3$ since the results were similar for all the DNNs evaluated. As expected, we empirically confirmed that also for this particular situation, \texttt{$\epsilon$-ProVe} returns a correct solution.

From the results of Tab\;\ref{tab:fullresults}, we can see that our approximation returns a tight, safe rate (mean of underestimation 1.27\%) for each model tested in at most half a minute, confirming the effectiveness and correctness of our approach.

\subsection{\#DNN-Verification on \texttt{$\epsilon$-ProVe} Result}

We performed an empirical experiment using an exact count method to confirm the safety probability guarantee of Theorem \ref{th_eProVe}. Specifically, theorem \ref{th_eProVe} proves that\texttt{$\epsilon$-ProVe} returns with probability $\alpha$ a set of safe areas ${\cal R}^{(\epsilon)}$, where for each hyperrectangle $r \in {\cal R}^{(\epsilon)}$ at most $(1-R)\cdot|r|$ points are not safe. This experiment aims to empirically verify this safety guarantee with a formal method. To this end, we rely on ProVe \cite{ProVe}, an exact count method that returns the violation rate, i.e., the portion of the area that presents violation points in a given domain of interest.

Due to scalability issues of exact counters, discussed in detail in the main paper, we only perform this evaluation on each of the $252$ hyperrectangles returned by \texttt{$\epsilon$-ProVe} for the $model\_2\_68$. The formal results are reported in Tab. \ref{tab:exact_eval}.

The exact count method used on each hyperrectangle $r$ returned by \texttt{$\epsilon$-ProVe} confirms the correctness of our approach. Since we set $R=99.5\%$ as a provable safe lower bound guaranteed, we expected at most a $0.05\%$ violation rate on each hyperrectangle $r_i$. Crucially, we notice that for only $2$ of $252$ hyperrectangles tested, we have a violation rate over the $0\%$, and this value is strictly less than $0.05\%$. Hence, also this experiment confirms the strong probabilistic guarantee presented in point 2. of Theorem \ref{th_eProVe}.

\end{document}